\pdfoutput=1
\documentclass[11pt]{article}

\usepackage{ACL}

\usepackage{times}
\usepackage{latexsym}
\usepackage{graphicx}
\usepackage{booktabs}
\usepackage{multirow}
\usepackage{amsmath}
\usepackage{amsfonts}
\usepackage{subcaption}
\usepackage{tikz}
\usepackage{float}
\usepackage{tcolorbox}
\usepackage{float}
\usepackage{soul}
\usepackage{algorithm}
\usepackage{algpseudocode}
\usepackage{amssymb}
\usepackage[linguistics]{forest}
\usepackage{caption}
\usepackage{titlesec}
\usepackage[T1]{fontenc}
\usepackage[utf8]{inputenc}

\usepackage{microtype}

\usepackage{inconsolata}

\title{Learning Disentangled Semantic Spaces of Explanations \\ via Invertible Neural Networks}

\author{Yingji Zhang$^{1\dagger}$,~ Danilo S. Carvalho$^{1,3}$,~ Andr\'{e} Freitas$^{1,2,3}$ \\
  $^{1}$ Department of Computer Science, University of Manchester, United Kingdom\\
  $^{2}$ Idiap Research Institute, Switzerland\\
  $^{3}$ National Biomarker Centre, CRUK-MI, Univ. of Manchester, United Kingdom\\
  \texttt{\{firstname.lastname\}@[postgrad.]$^{\dagger}$manchester.ac.uk}}

\begin{document}
\maketitle
\begin{abstract}
Disentangled latent spaces usually have better semantic separability and geometrical properties, which leads to better interpretability and more controllable data generation. While this has been well investigated in Computer Vision, in tasks such as image disentanglement, in the NLP domain sentence disentanglement is still comparatively under-investigated. Most previous work have concentrated on disentangling task-specific generative factors, such as sentiment, within the context of style transfer. In this work, we focus on a more general form of sentence disentanglement, targeting the localised modification and control of more general sentence semantic features. To achieve this, we contribute to a novel notion of sentence semantic disentanglement and introduce a flow-based invertible neural network (INN) mechanism integrated with a transformer-based language Autoencoder (AE) in order to deliver latent spaces with better separability properties. Experimental results demonstrate that the model can conform the distributed latent space into a better semantically disentangled sentence space, leading to improved language interpretability and controlled generation when compared to the recent state-of-the-art language VAE models.

\end{abstract}
\section{Introduction}

\begin{figure}[t]
    \centering
    \includegraphics[scale=0.83]{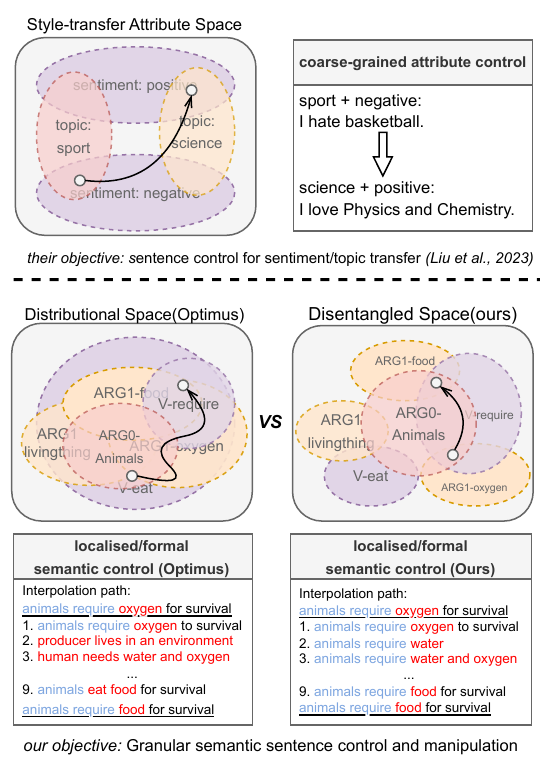}
    \caption{Top: attribute space geometry. Bottom: general semantic geometry, where left: distributional semantic space of Optimus \cite{li2020optimus}, right: our compositionality-induced semantic space where the geometrical location of sentence vectors can be located by the intersection of role-content clusters.}
    \label{fig:comparison}
\end{figure}

Most previous work on controlled text generation have concentrated on style transfer tasks: modifying sentences with regard to markers of sentiment, formality, affirmation/negation \cite{john2019disentangled,bao2019generating,hu2021causal,vasilakes-etal-2022-learning,gu-etal-2022-distributional,liu-etal-2023-composable,gu-etal-2023-controllable} (Figure \ref{fig:comparison} top). Disentanglement of language generative factors over Variational Autoencoder (VAE) spaces has been a key mechanism to deliver this type of generative control \cite{john2019disentangled,bao2019generating,vasilakes-etal-2022-learning}. 
%However, the investigation of disentanglement methods has been mainly contained in disentangling task-specific linguistic factors, especially in style transfer tasks. 
Recently, \citet{zhang2022} demonstrated that a more general form of semantic control can be achieved in the latent space of Optimus \cite{li2020optimus}, the first standard transformer-based VAE, where a BERT \cite{devlin2019bert} encoder and a GPT2 \cite{Radford2019LanguageMA} decoder are connected within a VAE bottleneck. Using representations of conceptually dense \textit{explanatory sentences} \cite{jansen2018worldtree}, they showed that sentences (e.g. \textit{animals require oxygen for survival}), can be represented within a space which can be organised around the associations between predicate, arguments and their associated token content: \textit{ARG0-animals} or \textit{VERB-require}, is geometrically resolved to a hypersolid over the latent space. Nevertheless, the ability to learn and control such separation is still limited as different semantic factors of the sentence are still overlapped and entangled in the latent space (e.g., \textit{V-eat} and \textit{V-require} in Figure \ref{fig:comparison} bottom left), indicating distributional sentence semantics cannot be currently localised and controlled from the perspective of formal semantics (i.e., \textit{predicate-argument structures}, \textit{compositionality}) \cite{marcus2003algebraic,Nefdt2020-NEFAPC,dankers-etal-2022-paradox}.

This work aims to improve the localisation and semantic control of latent sentence spaces, by delivering a model which can better separate and control syntactic-semantic features (e.g. predicate-argument) and their associated lexical semantics content. This type of representation can provide the foundation to shorten the gap between deep latent semantics and formal linguistic representations \cite{10.3115/1075218.1075283,banarescu2013abstract,mitchell2023we}, integrating the flexibility of distributional-neural models with the properties of linguistically grounded representations, facilitating both interpretability and generative control.

To deliver this type of semantic control within the distributional sentence space, following the methodological framework introduced by \cite{zhang2022}, we target on improving the semantic separability of sentences by focusing on explanatory sentences \footnote{The rationale for choosing explanatory sentences and their semantic details are provided in Appendix \ref{sec:train_detail}.}, rather than synthetic or style transfer datasets \cite{hupkes2020compositionality,yanaka2021sygns}, in which the \textit{semantic structure of sentences} can be isolated and controlled. Inspired by the work of \citet{esser2020disentangling}, we integrate a flow-based invertible neural network (INN) \cite{dinh2014nice} as a plug-in control component to learn the bijective transformation between the distributional hidden space of the transformer-based language autoencoder (BERT-GPT2) and the smooth Gaussian space of the INN (Figure \ref{fig:optimus_latent_space}). Specifically, we first pre-train an autoencoder (AE) to learn sentence representations from the transformers' latent spaces. Then, we freeze the AE weights and train the INN to map the AE representations to a Gaussian space. Since INN models define a bijective transformation, we can control the autoencoder generation by manipulating the INN latent spaces, which is more efficient and significantly less resource intensive than re-training a language AE end-to-end.

% \begin{figure}[t] 
%     \centering
%     \includegraphics[scale=0.28]{figs/verb_cluster.pdf}
%     \caption{t-SNE plot for VERB-$\{$is, are, cause, require$\}$ clusters (blue: are, green: cause, red: is, purple: require) (left: Optimus, right: our supervised INN).}
%     \label{fig:verb_example}
% \end{figure}
% We propose a way to separate the role-content cluster of explanatory sentences in the latent space with the target of improving latent semantic geometry. 

More importantly, we propose a supervised training strategy within the INN setting to learn a latent space with improved semantic separability, namely: the \textit{semantic role-content pairs} and their associated clusters can be better separated over the latent space modelled by the INN (Section \ref{sec:cluster_sup}). In this case, we can improve localised control over the decoding process due to the reduction of overlapping (ambiguous) regions. \textit{A more separable and geometrically consistent sentence space} can be then operated over to improve the generative control with support of geometric operators, such as interpolation \cite{bowman2016generating} (Section \ref{sec:guide_interpolate}). The contributions of this work are summarised below:

\textbf{1.} We approach sentence disentanglement and generation control from the point of view of \textit{Argument Structure Theory (AST)}, bridging latent space features with a canonical, linguistics-informed, semantic representation of sentences. \textbf{2.} We find that integrating a flow-based INN mechanism into a transformer-based language-AE architecture is an effective mechanism for transforming the hidden space of the autoencoder into a smooth Gaussian latent space for representing sentences. \textbf{3.}  We propose a supervised training strategy for INNs to learn a controllable semantic space with higher disentanglement and separability of semantic features, when compared to previous work. \textbf{4.} Using this mechanism, we systematically employ geometrical data augmentation strategies to assist on sentence representation disentanglement. 

Interpreting and controlling sentence generation from the perspective of the geometric manipulation of the latent space is still largely unexplored within NLP. To the best of our knowledge, this is the first work which focuses on the introduction of invertible NN-based mechanisms to support latent spaces with better separated argument structure/semantic features, allowing for a more universal form of sentence generation control.

\section{Preliminaries} \label{sec:disentang}
In this section, we first introduce the formal semantics and define the sentence representation model based on Argument Structure Theory (AST), linking with the associated disentanglement/generative factors and then proceed with the description of the proposed flow-based INN mechanism.

\paragraph{Controllability and interpretation in formal semantics.} Formal semantics, which provides a canonical, granular, and rigid representation, have been investigated for thousands of years, with well established theoretical frameworks such as Montague Semantics \cite{dowty2012introduction}, Davidsonian Semantics \cite{davidson1967logical}, Neo-Davidsonian Semantics \cite{lasersohn2016semantics}, etc. One typical characteristic of these formal semantics is the localisation or composition property. For example, in the sentence, \textit{animals require oxygen for survival}, the words are functionally combined into sentence semantics: 
\begin{equation} \label{eq:1}
    \lambda x (\text{animals}(x) \rightarrow \text{require}(x, \text{oxygen}))
\end{equation}
Where \( x \) is the variable representing any entity within a logical structure. In this case, we can localise the sentence semantics by replacing $x$ with \textit{birds}, \textit{fishes}, etc. This localised process indicates the interpretation in Cognitive Science \cite{smolensky2006harmony,lees1957syntactic}.
Disentanglement \cite{bengio2013deep} can potentially provide such localisation in the context of distributional latent representations, which has been widely investigated to localise image generation \cite{esser2020disentangling,jeon2019efficient, liu2021smoothing}. Therefore, we link several key notions—\textit{disentanglement}, \textit{formal semantics}, and \textit{localization}—to investigate formal control and interpretability in language models.

\paragraph{Sentence semantic disentanglement.} AST \cite{jackendoff1992semantic, levin1993english, rappaport2008english} provides a model for representing sentence structure and meaning of sentences in terms of the interface between the their syntactic structure and the associated semantic roles of the arguments within those sentences. It delineates how verbs define the organisation of their associated arguments and the reflection of this organisation in a sentence's syntactic realisation. AST abstracts sentences as predicate-argument structures, where the predicate $p$ (associated with the verb) has a set of associated arguments $arg_i$, where each argument has an associated positional component $i$ and a thematic/semantic roles $r_i$, the latter categorising the semantic functions of arguments in relation to the verb (e.g. agent, patient, theme, instrument). In the context of this work, the AST predicate-argument representation is associated with a lexical-semantic representation of the content $c_i$ of the term $t_i$. 

In this work, we simplify and particularise the relationship between the argument structure and the distributional lexical semantic representation as a \textit{role-content} relation, where the structural syntactic/semantic relationship is defined by its shallow semantics, i.e. as the composition of the content of the terms, their position in the predicate-argument (PArg) structure ($arg_i$) and their semantic roles (SRs) ($r_i$: $pred$, $arg$). Therefore, this work uses the notion of sentence semantic disentanglement as the cluster separation of the content under the PArg/SRs structure (the corresponding \textit{role} in role-content), aiming to induce a latent space which geometrically encodes the AST structure, better disentangling and separating role-content clusters. 
% (Figure~\ref{fig:comparison}).
%In view of the \textit{principle of compositionality} (Frege's principle), sentence semantics can be seen as consisting of word-level semantics, which can be jointly represented by word content and its corresponding syntactic/semantic role \cite{pelletier1994principle}. In this work, we simplify and particularise this relationship as (\textit{role-content} pair), where the structural syntactic/semantic relationship is defined by its shallow semantics, i.e. as the composition of the content of tokens and their semantic role labels (SRLs). Therefore, this work uses the notion of sentence semantic disentanglement as the cluster separation of the content under SRLs, where ``all entangled" dimensions contribute to the ``disentangled/separated geometrical location" of role-content clusters. Such a notion can better utilize the representation capability of the current distributional sentence representation in the disentanglement task rather than following the notion of feature-dimension binding, where each feature of data refers to an independent latent dimension in image disentanglement.

% $\underbrace{animals}_{ARG0}~\underbrace{require}_{PRED}~\underbrace{oxygen}_{ARG1}~\underbrace{for~survival}_{ARGM-PRP}$\\

\begin{table}[ht!]
\centering
\resizebox{7cm}{!}{
\begin{forest}
for tree={
    if n children=0{
      inner sep=1pt,
      align=center,
      base=top,
      s sep=0mm,
      l sep=100mm,
      minimum size=1em,
      tier=terminal
    }{},
  }
  [S [NP [NN [ \textit{ARG0 (Agent)} [\textit{animals}] ] ]] [VP [VBZ [ \textit{PRED} [\textit{require}]] ] [NP [ NP [\textit{ARG1 (Patient)} [\textit{oxygen}] ] ] 
  [PP [ARGM-PRP [\textit{PRP} [\textit{for}] ] 
  [\textit{PRP} [\textit{survival}] ]
  ]]]]]
\end{forest}
}
\end{table}
% As the transformer-based encoder can capture the complex structural information, such as syntax \cite{jawahar-etal-2019-bert}
%According to \textit{Compositional Distributional Semantics} \cite{clark2008compositional}, after encoding in latent space, the semantics of each sentence representation $s$ can be described as:
Formally, a sentence $s$ (e.g., see above) consists of a sequence PArgs/SRs and word content associations. Upon encoding in latent space, this can be described as:
\[
sem(s) = \underbrace{t_1({c_1}, {r_1})}_{i.e., ARG0-animals} \oplus \dots \oplus \underbrace{t_i({c_i}, {r_i})}_{PRP-survival}
\]
where $t_i({c_i}, {r_i})=c_i \otimes r_i$ represents the semantics of term $t_i$ with content $c_i$ (i.e., \textit{animals}) and SRL $r_i$ (i.e., \textit{ARG0}) in context $s$, $\otimes$: connects the meanings of words with their roles, using the compositional-distributional semantics notation of \cite{smolensky2006harmonic,clark2008compositional}. $\oplus$: connects the lexical semantics (word content + structural role) to form the sentence semantics. This work applies distinct symbols aiming to emphasise the disentanglement aspects associated with the AST structure. If the sentence representation can be semantically disentangled under $\oplus$, the $sem(s)$ can be decomposed into: 
$$sem(s) = \{ t_1({c_1}, {r_1})\} \oplus \dots \oplus \{t_i({c_{i}}, {r_{i}}) \}$$
% &= \{w_i({c_i}, {r_i})\} \oplus \{ w_1({c_1}, {r_1}) \} \\
%     &\oplus \{w_2({c_{2}}, {r_{2}}) \oplus \dots \oplus w_1({c_{i-1}}, {r_{i-1}}) \}
where each set represents a specific role-content cluster resolved to a hypersolid over the latent space, in this case, given a set of $N$ sentences within the same predicate cluster $t({c}, {r})$ (i.e., \textit{V-require}) but different $sem(s)$, those sentence vectors can represent $t({c}, {r})$ features independently of other features (i.e., \textit{ARG0-animals}), forming the $t({c}, {r})$ cluster:
$$\{sem(s_1), ..., sem(s_N)\} = \{t({c}, {r})\}_{\times N} \oplus \{... \}$$ 
Therefore, we can evaluate the disentanglement (i.e., \textit{natural clustering property} \cite{bengio2013deep}) of sentence semantics by evaluating the density within $\{t({c}, {r})\}$ set(cluster) (classifier recall) and the separation between different $\{t({c}, {r})\}$ set(clusters) (classifier accuracy) with downstream classifiers based on the \textit{manifold hypothesis for classification} \cite{Rifai2011TheMT}, rather than disentanglement metrics, which usually calculate the separation between latent dimensions, commonly used in the image domain \cite{higgins2016beta, kim2018disentangling, chen2018isolating, ridgeway2018learning}. Next, we will introduce the INN-based mechanism which is used to learn this semantically disentangled space.
% \begin{figure}[t]
%     \centering
%     \includegraphics[scale=0.73]{figs/cluster_illustrate_new.pdf}
%     \caption{In semantically disentangled space, sentence vectors, $\circledast$, can be located by the intersection of role-content clusters.}
%     \label{fig:cluster_sep_illustrate}
% \end{figure}

\paragraph{Invertible Neural Networks (INNs).}
Flow-based INNs \cite{dinh2014nice,dinh2016density} are a class of neural networks that models the bijective mapping between the observation distribution $p(x)$ and the latent distribution $p(z)$. We use $T$ to represent the forward mapping (from $p(x)$ to $p(z)$) and $T'$ to represent the backward mapping (from $p(z)$ to $p(x)$), respectively. Unlike VAEs that approximate the prior distribution to multivariate Gaussian distributions, INNs exactly use multivariate Gaussian distributions. These are trained by the following objective function: $\mathcal{L} = \nonumber
- \mathbb{E}_{x \sim p(x)} \Big[ T(x) \Big]^2 + \log \left| T'(x) \right|$ where $T(x)$ learns the transformation from $x$ to $z \sim N(0, 1)$. $\left| T'(x) \right|$ is the determinant of the Jacobian for $T(x)$, which indicates the extent in which the transformation locally expands or contracts the space. The term $\log \left| T'(x) \right|$ ensures the integration of the probability density function to be one. The forward and reversed mapping can be implemented via the \textit{coupling} layer \cite{dinh2014nice, kingma2018glow}.

The rationale for choosing flow-based INNs lies on the fact that they learn the bijective transformation between the latent and observed spaces, which can be used to guide the autoencoder generation by manipulating the INN latent space, which is more efficient and has lower computational demand than re-training a language VAE. Besides, flow-based INNs that learn the prior distribution (i.e., Gaussian) exactly, can theoretically prevent the information loss from variational inference (ELBO) where the prior is approximated from posterior $P(z|x)$.

% As for forward mapping, its basic form can be described as follows:
% $$
% T(x) = \begin{cases} 
% z_1 = x_1 \\
% z_2 = x_2 ~opr~ m_{\theta} (x_1)
% \end{cases}
% $$
% Where $[x_1;x_2]=\text{split}(x)$, $[z_1; z_2]=\text{split}(z)$, $m_{\theta}$ is any kind of network. The reversed mapping can be obtained:
% $$
% T'(z) = \begin{cases} 
% x_1 = z_1 \\
% x_2 = z_2 ~opr'~ m_{\theta} (z_1)
% \end{cases}
% $$
% Where ($opr$, $opr'$) are symmetrical mathematical operators in INNs, such as ($+$, $-$) or ($\odot$, $\div$).

\section{Proposed Approach} \label{sec:latent_props}
\begin{figure}[t]
    \centering
    \includegraphics[width=\columnwidth]{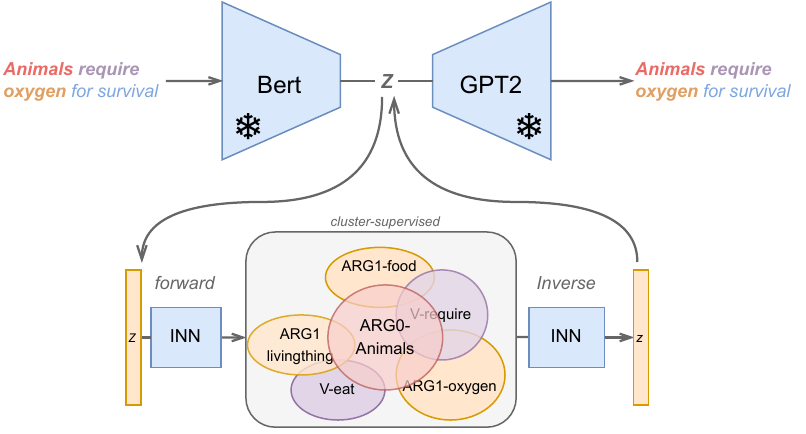}
    \caption{Transforming the representations of explanatory sentences from a language autoencoder (BERT-GPT2), into asemantically separable latent space with the support of the INN mechanism, where a sentence representation can be decomposed into a predicate-argument-level semantics (role-content).}
    \label{fig:optimus_latent_space}
\end{figure}

We encode each sentence $x$ with a frozen autoencoder (e.g., Bert-GPT2) and consider its sentence representation $E(x)$ as the input of INNs (Figure \ref{fig:optimus_latent_space}). We propose two training strategies to map the hidden representations into the Gaussian space.
\subsection{Training Strategy}
\paragraph{Unsupervised INN.} Firstly, we train the INN-based model unsupervised, which minimises the negative log-likelihood of the marginal distribution of latent representation $z=E(x)$:\\ 
$$\mathcal{L}_\text{unsup} = 
- \mathbb{E}_{x \sim p(x)} \Big[ T(E(x)) \Big]^2 + \log \left| T'(E(x)) \right| \nonumber$$
As the minimisation leads to a bijective mapping between the distributed representation and the disentangled latent representation (multivariate Gaussian space), it allows for a more semantically consistent representation of the geometric (role-content) clustering properties of its latent space, allowing for a more consistent traversal and interpolation \cite{li2020optimus} over the sentence space (Figure~\ref{fig:comparison}).

% \noindent
% \paragraph{Contrastive supervised INN} 
% Inspired by the work \citep{he2017deep} in the field of computer vision, we secondly train INN by minimizing the posterior distribution of semantic role $F$, such as ARG0, predicate, and ARG1, based on sentence pairs $(x^a, x^b | F)$ where $x^a$ and $x^b$ share one specific semantic role $F$. For example, given a sentence pair $x^a$ is "\textit{\textcolor{blue}{animal} require food for survival}" and $x^b$ is "\textit{\textcolor{blue}{animal} eat living things}", both of them hold the same content of ARG0 - \textit{animal}. In this case, the model learns the mapping function which makes them close to each other over the latent space. 

% In more detail, for a given training pair $(x^a, x^b) \sim p(x^a, x^b|F)$, we minimize the posterior distribution $p(z^a|x^a)$ where $p(z^a|x^a) \sim N(0, 1)$. Simultaneously, we also minimize $p(z^b|x^b)$ where $p(z^b|x^b) \sim N(\sigma z^a, (1-\sigma^2)1)$. $\sigma$ is a correlation hyper-parameter for controlling the similarity. The final objective function can be described as follow:
% \begin{equation} \label{contrastive_sup_inn}
% \begin{split}
% \mathcal{L}_\text{sup} =
% & - \mathbb{E}_{x^a \sim p(x^a, x^b)} \Big[ T(E(x^a)) \Big]^2 \\
% & - \beta^{-1} \log \left| T'(E(x^a)) \right| \\ 
% & - \mathbb{E}_{x^b \sim p(x^a, x^b)} \frac{\Big[ T(E(x^b)) - \sigma T(E(x^a)) \Big]^2}{1-\sigma^2} \\
% & - \beta^{-1} \log \left| T'(E(x^b)) \right|
% \end{split}
% \end{equation}
\paragraph{Cluster-supervised INN.} According to the findings of \cite{zhang2022}, the content of the predicate-argument structure/semantic roles can be disentangled over the latent space approximated to multivariate Gaussian learned using the Language VAE setting. Using the same foundation, we train the INN component to learn the embeddings, by minimising the distance between points in the same role-content regions and maximising the distance between points in different regions, based on the explanation embeddings and their corresponding central point from the language autoencoder model. For example, given a sentence "\textit{\textbf {animals} require food for survival}" and its central vector of \textit{ARG0-animals}, the training moves the sentence representation closer to the \textit{ARG0-animals} region centre in the INN latent space. Specifically, during the calculation of the posterior, we replace the mean and variance of the standard Gaussian distribution by the centre point of its cluster and a hyper-parameter, which should be less than one, respectively. In this case, each role-content cluster in the latent space will be mapped to a space where each cluster will have its embeddings more densely and regularly distributed around its centre. The objective function can be described as follows:
\vspace{-0.2cm}
\begin{equation}
\begin{split}
\mathcal{L}_\text{sup} = 
& - \mathbb{E}_{x \sim p_{cluster}(x)} \frac{\Big[ T(E(x)) - \mu_{cluster} \Big]^2}{{1-\sigma^2}} \\  \nonumber
& +  \log \left| T'(E(x)) \right|
\end{split}
\end{equation}
\noindent where $T(E(x))$ learns the transformation from $x$ to $z \sim N(\mu_{cluster}, 1-\sigma^2)$. $\sigma^2$ is a parameter which can be empirically determined (in this particular context the optimal value was found to be 0.6). Additional details are provided in Appendix \ref{sec:train_detail}. 
% To better represent the common features within the same role-content and different features between distinct role-content clusters, more training sentences are needed in those clusters. Therefore, 
\subsection{Geometrical Data Augmentation}
Data augmentation, which captures and augments a common or distinct feature across different samples, has been considered a common technique to assist disentanglement, such as in Graph \cite{li2021disentangled} and Image \cite{liu2022learning} representations, but is still limited in the context of sentence generation. In this work, we consider the vector arithmetic and traversal operators as a systematic mechanism to support data augmentation (via semantically controlled sentence generation) for each role-content cluster, described as follows: 
\begin{equation}
    \begin{aligned}
        &(1)\quad \text{v} = average(E'(x_i), E'(x_j)) \\ \nonumber
        &(2)\quad \text{v}_{neighbour} = \text{v}[i] \sim N(0, 1)_{\forall i \in \{0,..,size(\text{v})\}} \\
        &(3)\quad x_{new} = D'(\text{v}_{neighbour})
    \end{aligned}
    \label{eq:data_augmentation}
\end{equation}
\noindent where $x_k \in S$ (original corpus), $E'$ and $D'$ are the encoder and decoder of Optimus fine-tuned over $S$. $average$ operation aims to modify the sentence while maintaining the target role-content common to both $x_i$ and $x_j$ \cite{zhang2022}. The term $\text{v}[i] \sim N(0, 1)$ is introduced to resample each dimension of $\text{v}$ in the latent space (i.e., traverse its neighbour) and $x_{new} = D'(\text{v}_{neighbour})$ generates a new sentence. Finally, we only keep the sentences holding the target role-content, where the PArgs/SRs of $x$ are annotated via the \textit{AllenNLP}  \cite{gardner2018allennlp} semantic role labeller. Table \ref{tab:aug_data} lists randomly selected examples from augmented explanations. Full details and the supporting ablation study are provided in Appendices \ref{sec:train_detail} and \ref{sec:ablation}.
% \begin{table}[t]
%     \small
%     \centering
%       \resizebox{7.6cm}{!}{
%     \begin{tabular}{|c|cc|}
%         \hline
%         Task & Num data. & Avg. length \\ \hline
%         Explanation Generation & 11430 & 8.65 \\
%       	Natural Language Inference & 5134 & - \\ \hline
%     \end{tabular}
%     }
%     \caption{Statistics from explanations datasets.} \label{tab:stats_data}
% \end{table}

\begin{table}[ht!]
    \small
    \centering
      \resizebox{7.6cm}{!}{
    \begin{tabular}{ll}
        \toprule
        Role-content & Augmented sentences \\ \hline
        \multirow{3}{*}{ARG0-animal} 
        & \textcolor{blue}{an animal} requires energy to move \\
        % & \textcolor{blue}{animals} produce offspring \\
        % & \textcolor{blue}{some adult animals} lay eggs  \\
        & \textcolor{blue}{an animal} requires shelter \\
        & \textcolor{blue}{an animal} can use its body to breathe \\ \hline
        \multirow{3}{*}{ARG0-human} 
        % & \textcolor{blue}{humans} travel sometimes \\
        & \textcolor{blue}{humans} usually use gasoline \\
        % & \textcolor{blue}{humans} sometimes endanger themselves \\
        & \textcolor{blue}{humans} use coal to make food \\
        & \textcolor{blue}{humans} depend on pollinators for survival \\ \hline
        \multirow{3}{*}{PRED-are} 
        & wheels \textcolor{blue}{are} a part of a car \\
        & lenses \textcolor{blue}{are} a part of eyeglasses \\
        % & toxic chemicals \textcolor{blue}{are} poisonous \\/
        % & green plants \textcolor{blue}{are} a source of food for animals \\
        & copper and zinc \textcolor{blue}{are} two metals \\ \hline
        \multirow{3}{*}{PRED-mean} 
        & summit \textcolor{blue}{mean} the top of the mountain \\
        & colder \textcolor{blue}{mean} a decrease in heat energy \\
        % & helping \textcolor{blue}{mean} something can be done better \\
        % & cleaner \textcolor{blue}{mean} ( less ; lower ) in pollutants \\
        & friction \textcolor{blue}{mean} the product of a physical change \\ \toprule
        
    \end{tabular}
    }
    \caption{Augmented explanations. We also provide more examples in Table \ref{tab:aug_data_app} for qualitative evaluation.} \label{tab:aug_data}
\end{table}

\section{Experiments} \label{sec:empirical}
For the experiments, we start by focusing on the effect of the supervised INN mechanism to examine its impact on the sentence semantic separability of the distributional latent space defined in Section \ref{sec:disentang} (detailed in Section \ref{sec:cluster_sup}). Next, we examine the localised semantic generation control enabled by such semantic separability via latent interpolation (Section \ref{sec:guide_interpolate}). Further details of the AutoEncoder model and dataset are provided in Appendix \ref{sec:train_detail}.

\subsection{Disentanglement Encoding Evaluation} \label{sec:cluster_sup}
We examine the latent space separability (i.e., \textit{natural clustering property} \cite{bengio2013deep}) of our supervision approach on different predicate-argument/semantic roles. In the context of this work, the thematic roles' labels are not referred to control the generation. Instead, we use the predicate argument position markers, e.g. including \textit{ARG0}, \textit{ARG1}, \textit{PRED(V)}, where each category has \textit{a)} four possible word contents ($c_i$), or \textit{b)} the same content (i.e., \textit{animal}) with different argument/roles, including \textit{ARG0,1,2}. We provide the reconstructed examples of INNs in Table \ref{tab:rec_explain}.

\begin{figure}[ht!]
\centering
    \includegraphics[scale=0.15]{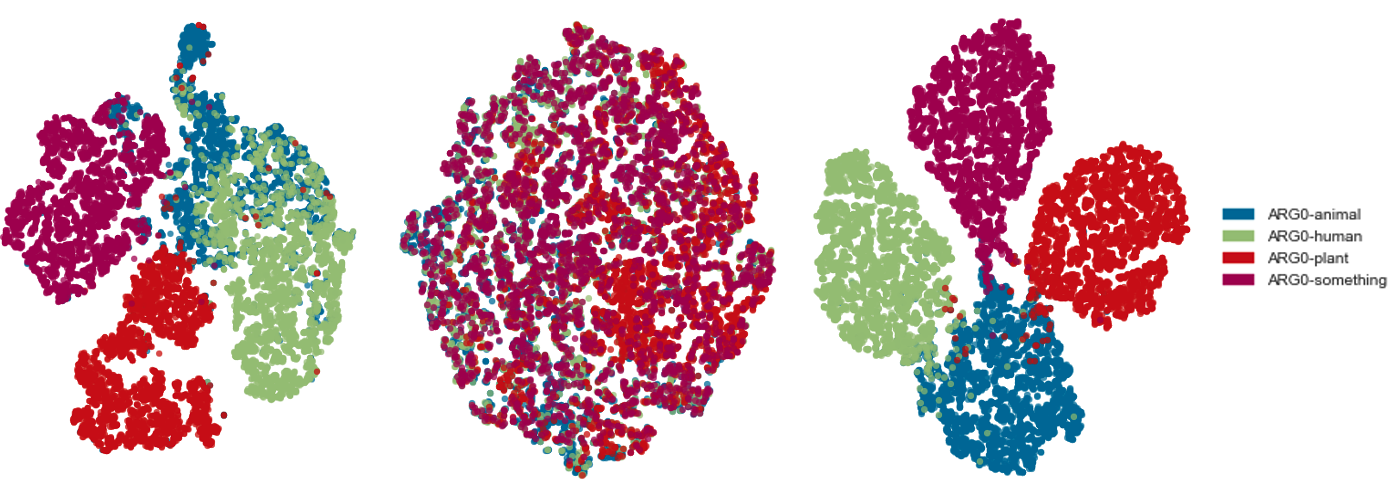}
    \caption{ARG0: t-SNE plot, different colour represents different content regions (blue: animal, green: human, red: plant, purple: something) (left: Optimus, middle: unsupervised, right: cluster supervised), same order for remaining visualizations. We also provide the PCA plot in Figure \ref{fig:a0_pca}, both visualization shows that supervised embeddings concentrate on the respective cluster center.}
    \label{fig:a0_sup}
\end{figure}

\paragraph{Disentanglement between \textit{ARG0} clusters.} For \textit{ARG0}, we choose \textit{human}, \textit{animal}, \textit{plant}, and \textit{something} due to having the highest frequency in the original dataset, and evaluate model performance from two directions, including forward and backward mapping. Within forward mapping, we assess the disentanglement of the latent space of the INN model from two perspectives (visualisation and classification metrics). Figure \ref{fig:a0_sup} displays the distributions of four role-content clusters over the latent space. As we can observe, after the cluster-supervised training strategy, the embeddings are more concentrated at the center of their cluster, and there is a clear boundary between clusters, indicating a better disentanglement when compared to Optimus and unsupervised INNs.
% It is also observable that there are low-density embedding regions at the transition (connection) between two clusters. We decode the middle datapoints between \textit{animal} and \textit{human} clusters and list them in Table \ref{tab:middle_examples}. From those examples, we can observe that such explanations are related to both \textit{animal} and \textit{human}. This result implies that the explanations may be geometrically represented in a similar way as they were originally designed in the WorldTree corpus (maximising lexical overlaps for pred-arg alignments within an explanation chain) for supporting multi-hop inference tasks.
% \input{tables/middle_explainations.tex}

We then quantitatively evaluate the disentanglement of ARG0-content clusters. We consider classification task metrics (\textit{accuracy}, \textit{precision}, \textit{recall}, \textit{f1}) as proxies for evaluating region separability, effectively testing cluster membership across different clusters. We choose a non-parametric downstream classifier (i.e., kNN) to quantitatively evaluate the separation of clusters and parametric downstream classifiers, including Naive Bayes (NB) and Support Vector Machine (SVM), to assess both separability and representation capability of latent sentence spaces \cite{Rifai2011TheMT, conneau-etal-2018-cram}. The configuration of the downstream classifiers are detailed in Appendix \ref{sec:train_detail}.
% % \begin{center}
% \begin{table}[ht]
% % \scriptsize
% \small
% \centering
% \setlength\tabcolsep{2.5pt}
% \resizebox{7.8cm}{!}{
% % \renewcommand\arraystretch{1.3}
% \begin{tabular}{cccccc}
% \toprule
% \multicolumn{6}{c}{ARG0: disentanglement proxy metrics} \\ \hline

% classifier & train & accuracy & precision & recall  & f1 score \\ \hline
% \multirow{3}{*}{KNN} & O & 0.983 & 0.983 & 0.983  & 0.983 \\
%  & U & 0.972 & 0.972 & 0.972  & 0.972 \\
%  & C & \textbf{0.986} & \textbf{0.986} & \textbf{0.986}  & \textbf{0.986} \\ \hline
 
% \multirow{3}{*}{NB} & O & 0.936 & 0.936 & 0.936  & 0.936 \\
% & U & 0.961 & 0.961 & 0.961  & 0.961 \\
%  & C & \textbf{0.979} & \textbf{0.979} & \textbf{0.979}  & \textbf{0.979} \\ \hline
 
% \multirow{3}{*}{SVM} & O & 0.979 & 0.979 & 0.979  & 0.979 \\
%  & U & 0.975 & 0.975 & 0.975  & 0.975 \\
%  & C & \textbf{0.981} & \textbf{0.981} & \textbf{0.981}  & \textbf{0.981} \\ \toprule
% \end{tabular}
% }

% \caption{Disentanglement of ARG0 between Optimus (O), unsupervised INN (U), and cluster-supervised INN (C) where KNN: k-neighbours, NB: naive bayes, SVM: support vector machine. The abbreviations are the same for the remaining tables. Cluster supervision displays consistent improvement with different classifiers.} \label{tab:arg0_exp}
% \end{table}
% % \end{center}

% \begin{center}
\begin{table}[ht]
% \scriptsize
\small
\centering
\setlength\tabcolsep{2.5pt}
\resizebox{7.8cm}{!}{
\begin{tabular}{cccccc}
\toprule
\multicolumn{6}{c}{ARG0: disentanglement proxy metrics} \\ \hline

classifier & train & accuracy & precision & recall  & f1 score \\ \hline
\multirow{3}{*}{KNN} & O & 0.972 & 0.973 & 0.972  & 0.972 \\
 & U & 0.938 & 0.938 & 0.938  & 0.938 \\
 & C & \textbf{0.979} & \textbf{0.979} & \textbf{0.979}  & \textbf{0.979} \\ \hline
 
\multirow{3}{*}{NB} & O & 0.934 & 0.934 & 0.933  & 0.933 \\
& U & 0.958 & 0.958 & 0.958  & 0.958 \\
 & C & \textbf{0.978} & \textbf{0.978} & \textbf{0.978}  & \textbf{0.978} \\ \hline
 
\multirow{3}{*}{SVM} & O & 0.970 & 0.970 & 0.970  & 0.970 \\
 & U & 0.972 & 0.972 & 0.972  & 0.972 \\
 & C & \textbf{0.980} & \textbf{0.980} & \textbf{0.980}  & \textbf{0.980} \\ \toprule
\end{tabular}
}

\caption{Disentanglement of ARG0 between Optimus (O), unsupervised INN (U), and cluster-supervised INN (C) where KNN: k-neighbours, NB: naive bayes, SVM: support vector machine. The abbreviations are the same for the remaining tables. Cluster supervision displays consistent improvement with different classifiers.} \label{tab:arg0_exp}
\end{table}
% \end{center}
As shown in table \ref{tab:arg0_exp}, all classifiers trained over supervised latent representations outperformed the unsupervised INN (U) and Optimus (O), indicating that the cluster-supervised approach leads to better disentanglement and representation. Moreover, (O) demonstrates superior performance compared to (U) for the KNN-based evaluation. However, it exhibits lower performance than (U) in NB and SVM. This suggests that the INN-AutoEncoder configuration can more effectively capture sentence semantics (from the point-of-view of AST+distributional content), in the context of a reconstruction task since the VAEs' training process is prone to experiencing posterior collapse.

\begin{figure}[ht!]
% \begin{center}
\centering
    \includegraphics[scale=0.15]{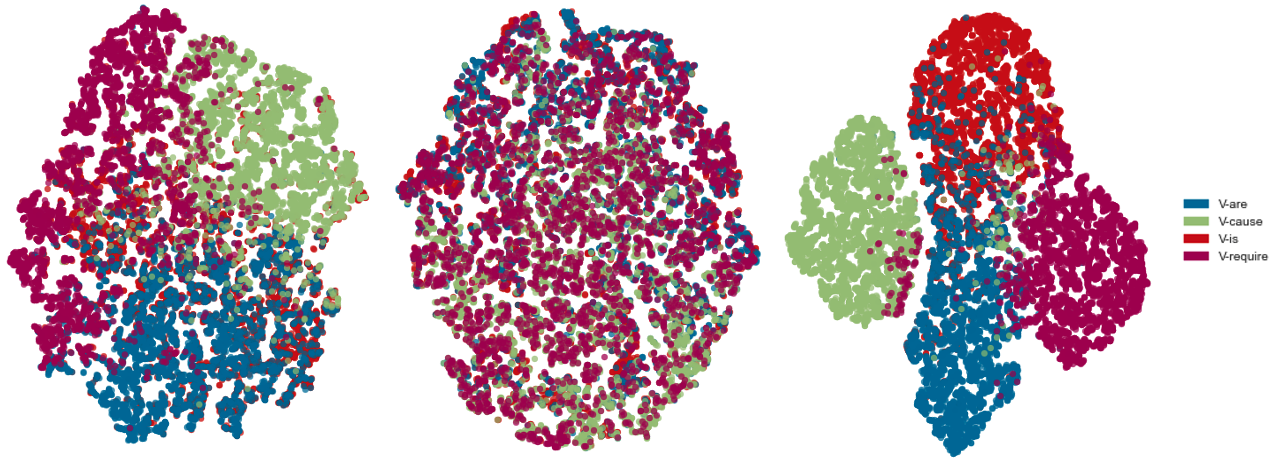}
    \caption{PRED: t-SNE plot (blue: are, green: cause, red: is, purple: require). PCA plot is in Figure \ref{fig:verb_pca}.}
    \label{fig:v_sup}
 % \end{center}
\end{figure}

As for the evaluation of the backward mapping, we calculate the ratio of generated sentences that hold the same role-content as the inputs (henceforth called the invertibility ratio). We randomly selected 100 embeddings as inputs and showed the corresponding ratios in Table \ref{tab:arg0_inv_exp}. We can observe that both unsupervised and supervised cases can achieve high invertibility ratios, indicating that the INN mechanism provides stable invertibility with or without cluster supervision.
%the means to control the sentence decoding step precisely by operating the vector over its transformed latent space. 
\begin{table}[ht!]
\small
\centering
\begin{tabular}{ccccc}
\toprule 
\multicolumn{5}{c}{ARG0: invertibility ratio (backward: $T'$)} \\ \hline
train & human & animal & plant & something \\ \hline
U & 0.980 & 0.890 & 0.990 & 1.000 \\
C & 1.000 & 0.860 & 0.990 & 0.950 \\ \toprule
\end{tabular}
\caption{Invertibility test for ARG0, Both INNs with AutoEncoder setup can achieve high ratios, indicating stable invertibility with or without cluster supervision.} \label{tab:arg0_inv_exp}
\end{table}
\paragraph{Disentanglement between \textit{PRED} clusters.} Next, we analyze the disentanglement between \textit{predicate (PRED)} clusters. As shown in Figure \ref{fig:v_sup}, although the disentanglement of PRED clusters is not as high as ARG0, the latent space with cluster supervision still performs better than both the unsupervised case and the Optimus model. In Table \ref{tab:v_exp}, the supervised INN model achieves better disentanglement, and both unsupervised and supervised could obtain a higher ratio. We also provide the experimental results of \textit{ARG1} disentanglement in Appendix \ref{sec:dis_pred}.

% % \begin{center}
% \begin{table}[ht]
% % \scriptsize
% \small
% \centering
% \setlength\tabcolsep{2.5pt}
% \resizebox{7.8cm}{!}{
% % \renewcommand\arraystretch{1.3}
% \begin{tabular}{cccccc}
% \toprule
% \multicolumn{6}{c}{PRED: disentanglement proxy metrics (forward: $T$)} \\ \hline

% classifier & train & accuracy & precision & recall  & f1 score \\ \hline
% \multirow{3}{*}{KNN}  & O & 0.964 & 0.964 & 0.964  & 0.964 \\
% & U & 0.959 & 0.959 & 0.959  & 0.959 \\ 
% & C & \textbf{0.972} & \textbf{0.972} & \textbf{0.972}  & \textbf{0.972} \\ \hline
 
% \multirow{3}{*}{NB}  & O & 0.923 & 0.923 & 0.923  & 0.923 \\
% & U & 0.927 & 0.927 & 0.927  & 0.927 \\ 
% & C & \textbf{0.951} & \textbf{0.951} & \textbf{0.951}  & \textbf{0.951}  \\ \hline
 
% \multirow{3}{*}{SVM}  & O & 0.956 & 0.956 & 0.956  & 0.956 \\
%  & U & 0.950 & 0.950 & 0.950  & 0.950 \\ 
%  & C & \textbf{0.958} & \textbf{0.958} & \textbf{0.958}  & \textbf{0.958} \\ \toprule
% \end{tabular}
% }
% \caption{Forward evaluation for predicate clusters, the invertibility ratio is provided in Table \ref{tab:v_invert_ratio}.} \label{tab:v_exp}
% \end{table}
% % \end{center}

\begin{figure}[ht]
\centering
    \includegraphics[scale=0.16]{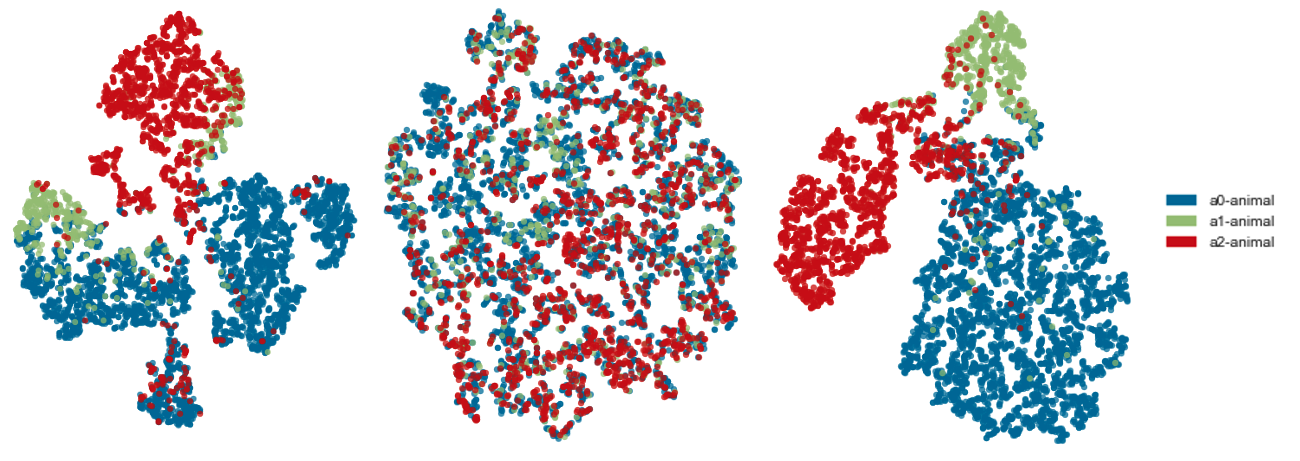}
    \caption{Animal: t-SNE plot (blue: ARG0-animal, green: ARG1-animal, red: ARG2-animal), PCA plot is provided in Figure \ref{fig:animal_pca}.}
    \label{fig:a012_animal}
\end{figure}

% \begin{center}
\begin{table}[ht]
% \scriptsize
\small
\centering
\setlength\tabcolsep{2.5pt}
\resizebox{7.8cm}{!}{
\begin{tabular}{cccccc}
\toprule
\multicolumn{6}{c}{PRED: disentanglement proxy metrics (forward: $T$)} \\ \hline

classifier & train & accuracy & precision & recall  & f1 score \\ \hline
\multirow{3}{*}{KNN}  & O & 0.911 & 0.914 & 0.910  & 0.911 \\
& U & 0.869 & 0.873 & 0.865  & 0.868 \\ 
& C & \textbf{0.922} & \textbf{0.927} & \textbf{0.918}  & \textbf{0.922} \\ \hline
 
\multirow{3}{*}{NB}  & O & 0.865 & 0.866 & 0.866  & 0.865 \\
& U & 0.873 & 0.874 & 0.871  & 0.872 \\ 
& C & \textbf{0.903} & \textbf{0.903} & \textbf{0.902}  & \textbf{0.903}  \\ \hline
 
\multirow{3}{*}{SVM}  & O & 0.902 & 0.902 & 0.903  & 0.902 \\
 & U & 0.905 & 0.906 & 0.902  & 0.904 \\ 
 & C & \textbf{0.910} & \textbf{0.912} & \textbf{0.909}  & \textbf{0.910} \\ \toprule
\end{tabular}
}
\caption{Forward evaluation for predicate clusters, the invertibility ratio and statistical significance test are provided in Table \ref{tab:v_invert_ratio} and \ref{tab:significant_test}.} \label{tab:v_exp}
\end{table}
% \end{center}
\paragraph{Disentanglement between \textit{ARG0,1,2} clusters.}
The experiments up to this point investigated the separation between the same pred-argument type but different content clusters. Next, we explore the separability of different pred-argument types with the same content. We thus focus on the \textit{animal} cluster, and investigate the disentanglement between \textit{ARG0-animal}, \textit{ARG1-animal}, and \textit{ARG2-animal}. As illustrated in Figure \ref{fig:a012_animal}, the animal clusters with different pred-argument types can be separated after cluster-supervised training, which indicates that the INN model can capture the difference between the same content with different pred-argument type in the case of similar topic, indicating the INN-based approach could jointly learn separable embeddings w.r.t. role-content and content alone. 

Table \ref{tab:animal_exp} and \ref{tab:animal_invert_ratio} show the disentanglement metrics and the invertibility ratio, respectively. Similarly to the previous experiment, the supervised case outperforms both the unsupervised and the Optimus models. Both INNs can achieve an invertibility ratio of at least 90\%. 

% We can observe that the INN-based model can generate explanations that hold the same role as its cluster, indicating that INN can separate the information of different semantic roles in similar contextual information.
% \begin{center}
\begin{table}[ht!]
% \scriptsize
\small
\centering
\setlength\tabcolsep{2.5pt}
% \resizebox{7.8cm}{!}{
\renewcommand\arraystretch{1}
\begin{tabular}{cccc}
\toprule
\multicolumn{4}{c}{Animal: disentanglement metrics (\textit{f1 score})} \\ \hline

train & KNN & NB & SVM  \\ \hline
O & 0.960 & 0.928 & 0.946 \\
U & 0.958 & 0.930 & 0.947 \\
C & \textbf{0.967} & \textbf{0.937} & \textbf{0.950} \\ \toprule
\end{tabular}
% }
\caption{Forward evaluation for Animal, we only show \textit{f1} since the same value across different metrics. Results indicate improved separation across role clusters.} \label{tab:animal_exp}
\end{table}

\subsection{Disentanglement Decoding Evaluation} \label{sec:guide_interpolate}
Finally, we evaluate the disentangled sentence geometry from the perspective of sentence generation. We specifically focus on linear interpolation as it can provide more efficient traversal between sentences and clusters than other traversal approaches (e.g., \textit{Ornstein-Uhlenbeck}), commonly used in the NLP domain \cite{li2020optimus} and in the evaluation of disentanglement \cite{bengio2013deep}.
\paragraph{Interpolation localisation.} Firstly, we evaluate the localisation of latent interpolation that interpolates a path $z_t = z_1 \cdot (1 - t) + z_2 \cdot t$  with $t$ increased from $0$ to $1$ by a step size of $0.1$, where $z_1$ and $z_2$ represent the latent representations of source and target sentences. As a result, $9$ sentences are generated on each interpolation step. On a latent space with better token-level role-content separation, given two sentences with the same role-content as endpoints, we can observe that the intermediate sentence can hold the same role-content during interpolation.
% In this experiment, we chose the unsupervised INN and Optimus as baselines\footnote{the standard transformer-based VAE(Optimus) with single sentence representation (i.e., the prior is standard Gaussian distribution). Some variant large VAEs, such as Della \cite{hu-etal-2022-fuse}, DPrior \cite{fang-etal-2022-controlled}, \cite{li-etal-2022-variational-autoencoder}, etc., were not included due to differing training objectives. Additionally, \citet{li2020optimus} have illustrated that Optimus can induce smoother interpolations than the Bert-GPT2 autoencoder. Therefore, we don't compare it in our work.} \paragraph{Interpolation localisation} 
% Compared with other latent space traversal approaches (e.g., \textit{Brownian} or \textit{Ornstein-Uhlenbeck} random walk), latent interpolation

In terms of qualitative evaluation, Table \ref{tab:guide_generation} provides the interpolation paths of cluster-supervised INN and Optimus, as for Optimus, we can observe that the intermediate explanations could transition smoothly from source to target for \textit{argument}.
% as for the unsupervised INN, we can observe that the intermediate explanations could transition smoothly from source to target for argument. E.g., moving from \textit{humans} to \textit{nonhumans} to \textit{marine animals} to \textit{animals}. 
However, the \textit{predicate} is more abruptly changed, indicating lower \textit{predicate-content} disentanglement (e.g., \textit{predicate-require} and \textit{predicate-eat}). Instead, the supervised INN can fix the \textit{predicate-require} during interpolation, indicating better separability between different predicate-content results in better generation control. More examples are provided in Table \ref{tab:guide_generation_app2} and \ref{tab:guide_generation_app1}.
\begin{table}[t]
\begin{tcolorbox}[fontupper=\small, fontlower=\small, title=interpolation localisation: \textit{predicate-require} ]
\underline{source: humans \textcolor{blue}{require} freshwater for survival}\\

Optimus: \\
1. humans \textcolor{blue}{require} water and food through fossil fuels \\
2. humans \textcolor{blue}{require} water for survival \\
3. humans \textcolor{red}{produce} small amounts of consumer food \\
4. human \textcolor{red}{has} a positive impact on a plant's survival \\
5. humans \textcolor{red}{convert} food into animal prey \\
6. humans \textcolor{red}{make} food for themselves by eating \\
7. animals \textcolor{blue}{require} food for survival \\
8. animals \textcolor{blue}{require} nutrients from the air \\
9. humans \textcolor{red}{eat} plants for food \\
10. animals \textcolor{blue}{require} food for survival \\

% Unsupervised INN: \\
% 1. nonhumans \textcolor{blue}{require} water to survive \\
% 2. marine animals \textcolor{blue}{require} food for survival  \\
% 3. animals \textcolor{red}{must breath} to survive  \\
% 4. animals \textcolor{blue}{require} water for survival \\
% 5. animals \textcolor{blue}{require} water from their ecosystems  \\
% 6. animals \textcolor{blue}{require} water for survival \\
% 7. animals \textcolor{red}{must eat} food for survival \\
% 8. animals \textcolor{blue}{require} food for survival \\
% 9. animals \textcolor{blue}{require} food for survival \\
% 10. animals \textcolor{blue}{require} food for survival \\

Cluster-supervised INN: \\
1. humans \textcolor{blue}{require} water for survival \\
2. nonhumans \textcolor{blue}{require} water for survival \\
3. animals \textcolor{blue}{require} water and food \\
4. animals \textcolor{blue}{require} water to survive \\
5. animals \textcolor{blue}{require} water to live \\
6. animals \textcolor{blue}{require} food for survival \\
7. animals \textcolor{blue}{require} food for survival \\
8. animals \textcolor{blue}{require} food for survival \\
9. animals \textcolor{blue}{require} food for survival \\
10. animals \textcolor{blue}{require} food to survive \\

% AutoEncoder
% 1. humans require water to survive \\
% 2. marine mammals require great amounts of water \\
% 3. animals require oxygen to survive  \\
% 4. animals require water for survival \\
% 5. animals \textcolor{red}{must eat} water to survive \\
% 6. animals require water and food  \\
% 7. animals require water for survival \\
% 8. animals \textcolor{red}{must eat} to survive  \\
% 9. animals require food for survival  \\
% 10. animals \textcolor{red}{must eat} food to survive \\
% unsupervised INN
% step 0, sent: nonhumans require water to survive
% step 1, sent: marine animals require food for survival
% step 2, sent: animals must breath to survive
% step 3, sent: animals require water for survival
% step 4, sent: animals require water from their ecosystems
% step 5, sent: animals require water for survival
% step 6, sent: animals must eat food for survival
% step 7, sent: animals require food for survival
% step 8, sent: animals require food for survival
% step 9, sent: animals require food for survival
% step 10, sent: animals require food for survival

\underline{target: animals \textcolor{blue}{require} food to survive}
\end{tcolorbox}
\caption{Interpolation examples, indicating the cluster-supervised INN can provide better localised/symbolic semantic control. We also report the interpolations of AutoEncoder and unsupervised INN in Table \ref{tab:guide_generation_app}.}
\label{tab:guide_generation}
\end{table}
\begin{figure}[ht!]
\centering
    \includegraphics[scale=0.35]{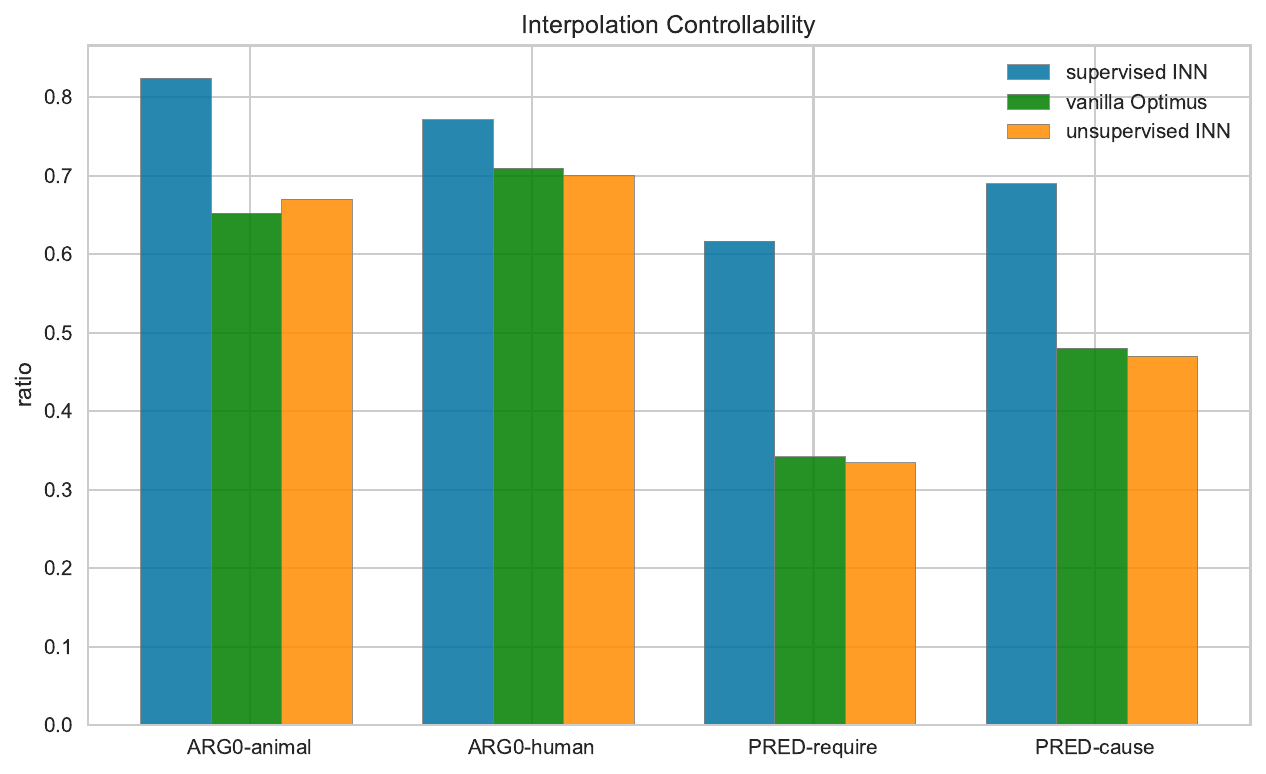}
    \caption{Interpolation control evaluation, we can observe that supervised INN with better semantic separability can lead to better localised semantic control.}
    \label{fig:interpolate_ratio}
\end{figure}
We then quantitatively evaluate the localisation of interpolation. We randomly select 200 sentence pairs from the dataset holding the same role-content and report the ratio of intermediate sentences with the same role-content as inputs. As illustrated in Figure \ref{fig:interpolate_ratio}, the intermediate sentences from the supervised INN can better hold the same role-content as inputs, especially for \textit{predicate} which usually has a lower effect on distributional sentence semantics \cite{zhang2022}, indicating that our supervision can lead to better latent space separability and generation control.

\paragraph{Interpolation smoothness.} Moreover, we quantitatively evaluate the latent space geometry via interpolation smoothness metrics (IS, \citet{Zhang2024ImprovingSC}), which calculates the ratio between the ideal semantic distance (i.e., the aligned semantic distance between source and target sentences) and the actual semantic distance (i.e., the sum of semantic distance between adjacent sentences during interpolation). A higher ratio indicates that the actual path aligns better with the ideal path, suggesting better semantic-geometric properties. The metric is defined as:
$$\text{IS} = \mathbb{E}_{(s_0, ..., s_T) \sim P} \frac{\delta(\text{align}(s_0, s_T))}{\sum^T_{t=0} \delta(\text{align}(s_t, s_{t+0.1}))}$$
where $s_0, ..., s_T$ is the sequence of sentences during interpolation, $\delta$ and $\text{align}$ are sentence similarity and alignment functions, respectively, which are performed via Word Mover’s Distance \cite{zhao-etal-2019-moverscore}.
We choose the standard language VAE baselines (i.e., the prior is the std. Gaussian distribution). Their implementation details are provided in Appendix \ref{sec:train_detail}. We randomly sample 200 sentence pairs and report the IS metric. As illustrated in Table \ref{tab:interpolation_smoothness}, our model can deliver smoother interpolations comparatively to the baselines, indicating semantic disentanglement can lead to better latent space geometry. 

\begin{table}[t]
\setlength\tabcolsep{2.5pt}
\small
\centering
\renewcommand\arraystretch{1}
\begin{tabular}{llll}
\toprule
Evaluation Metrics & avg IS$\uparrow$ & max IS$\uparrow$ & min IS$\uparrow$ \\ \hline
% \multicolumn{4}{c}{\textit{baselines without sentence botleneck}} \\ 
% Della \cite{hu-etal-2022-fuse} & \\
% T5VQVAE \cite{Zhang2024ImprovingSC} & \\ \hline
% \multicolumn{4}{c}{\textit{baselines with sentence bottleneck}} \\ 
DAE \cite{10.1145/1390156.1390294} & 0.144 & 0.330 & 0.055 \\
AAE \cite{makhzani2016adversarial} & 0.142 & 0.284 & 0.054 \\
LAAE\cite{rubenstein2018latent} & 0.172 & 0.347 & 0.056 \\ % \cite{rubenstein2018latent} 
DAAE \cite{shen2020educating} & 0.055 & 0.061 & 0.023 \\ % \citet{shen2020educating} 
$\beta$-VAE \cite{Higgins2016betaVAELB} & 0.198 & 0.379 & 0.041 \\ % \cite{Higgins2016betaVAELB}
% Optimus\cite{li2020optimus} & 0.1066 & 0.1555 & 0.0639 \\
% AutoEncoder (Bert-GPT2) & 0.1399 & 0.2166 & 0.0872 \\
% INN (U) (our) & 0.1315 & 0.1708 & 0.0791 \\
% INN (C) (our) & 0.1626 & 0.3474 & 0.1051 \\ 
AdaVAE \cite{tu2022adavae} & 0.085 & 0.105 & 0.050 \\
Della \cite{hu-etal-2022-fuse} & 0.253 & 0.416 & 0.155 \\ 
Optimus \cite{li2020optimus} & 0.220 & 0.525 & 0.130 \\
AutoEncoder (Bert-GPT2) & 0.259 & 0.585 & 0.165 \\
INN (U) (our) & 0.251 & 0.540 & 0.159 \\
INN (C) (our) & \underline{\textbf{0.282}} & \underline{\textbf{0.607}} & \underline{\textbf{0.206}} \\ 
\toprule
\end{tabular}
\caption{Geometrical examination via IS metric.} \label{tab:interpolation_smoothness}
\end{table}
\section{Related Work} \label{sec:related}
\paragraph{Sentence representation.} Sentence representations are usually trained in supervised \cite{conneau-etal-2017-supervised,Reimers2019SentenceBERTSE}, constrastive \cite{giorgi-etal-2021-declutr, yan-etal-2021-consert,chuang-etal-2022-diffcse}, or generation-oriented \cite{wang2021tsdae,wu-zhao-2022-sentence,chuang-etal-2022-diffcse} fashion. Recent work \cite{huang-etal-2023-bridging} explored the compositional sentence representation for improved explainability and generation. However, these works still lack the emphasis on the geometric interpretation and control of the underlying sentence space, which this work focused on.

\paragraph{Sentence disentanglement.} In the Natural Language Generation domain, most previous investigations explored the disentanglement between two specific linguistic perspectives, such as sentiment-content \cite{john2019disentangled,li-etal-2022-variational-autoencoder}, semantic-syntax \cite{bao2019generating}, and negation-uncertainty \cite{vasilakes-etal-2022-learning}, or syntactic disentanglement \cite{mercatali2021disentangling, felhi2022towards}. In this work, we provide a formal-geometrical lens, with the support of \textit{argument structures} as a sentence representation model, for sentence disentanglement targeting for localised semantic control. This work is the first integration of flow-based INN mechanisms to improve disentanglement, separation and semantic control of sentence spaces. 
% via latent sentence space geometry and propose flow-based INN autoencoders to achieve improved separation and control.

% \citet{mercatali2021disentangling} pioneered work on the use of disentanglement to control syntactic-level generative factors in sentence representations. More specialised architectures, such as the Attention-Driven Variational Autoencoder \cite{felhi2022towards}, were later introduced for learning disentangled syntactic latent spaces. Recently, \citet{carvalho2023learning} proposed a supervised training strategy for learning a disentangled representation of definitions by injecting the semantic role labelling inductive biases into the latent space. Before this, some studies explored the disentanglement of natural language on specific downstream tasks, such as text style transfer, paraphrasing, etc. \cite{bao2019generating, john2019disentangled}. We focus on latent sentence space geometry and propose~ flow-based INN autoencoders to achieve improved separation and control. 

\paragraph{INNs in NLP.} \citet{csahin2020two} concentrate on modelling morphological inflection and lemmatization tasks, utilizing INN to learn a bijective transformation between the word surface and its morphemes. \citet{li2020sentence} proposed BERT-flow, transforming sentences from a BERT sentence space to a standard Gaussian space. \citet{ding-gimpel-2021-flowprior} deployed flow-based INN to enrich VAE prior distribution, while \citet{gu-etal-2023-controllable} use flow mechanisms to control attributes in style transfer tasks. This work focused on semantic separability, geometrical operations and control over the distributed representation of sentences. Moreover, this work is the first to explore geometrical data augmentation to support semantic disentanglement.

\section{Conclusions and Future Work} \label{sec:concl}
This work focused on an INN-based mechanism to support better disentangled and separated latent sentence spaces over language autoencoders. By aligning the predicate-argument structure of sentences to the latent representations, we aimed to build a bridge between the formal and distributional semantics perspectives for sentence representation. We define the sentence semantic disentanglement from the perspective of formal semantics, aligning the predicate-argument structure to disentanglement and cluster separation properties, and exploiting the invertibility and bijection properties of INNs to facilitate such alignment. Experimental results indicate that the invertibility mechanisms can transform the distributed hidden space of an autoencoder into a latent space where AST-level syntactic and semantic transformations can be localised, interpolated and controlled. Secondly, we propose a supervision approach, which leads to an improved disentangled and separated space. This property can facilitate localised interpolation control. Lastly, we utilise these geometric properties to support a semantically controlled data augmentation to assist the disentanglement process.

Since our work connects distributional and formal semantics via disentanglement, one future direction is to explore the safety and control of the formal semantic properties of Large Language Models. Besides, recent work \cite{liu-etal-2023-composable} revealed that distinct factors can be composed by modelling the moving of latent vectors via ordinary differential equations, which can be adapted to sentence representations to deliver more complex sentence transformations within the latent space.

% Future work includes applying and evaluating our approach to downstream tasks such as natural language inference. 
% We intend to apply latent semantic separability on the NL Inference task where abstract-level retrieval is required.As this work explores the controllability of language from a less explored direction, we provide application guidelines for downstream tasks in Appendix \ref{sec:appl}

\section{Limitations} \label{sec:limit}
This work focused on the disentangled sentence representations geometry to deliver localised/semantic/formal semantic control. While this work is motivated by providing more localised distributed representations, which can positively impact the safety and coherence of generative models, few scoping observations need to be established: \textbf{1.} The specific safety guarantees of these models are not fully established. \textbf{2.} While the language autoencoder with unsupervised INN exhibit a distinct learning pattern with regard to semantic distribution, further understanding is required in terms of information bottleneck properties \cite{michael2018on} and on the semantic distribution of unsupervised INNs in language modelling tasks. \textbf{3.} Furthermore, this study exclusively focused on a corpus of sentences which are conceptually dense (\cite{https://doi.org/10.48550/arxiv.2104.08661}). The exploration of its performance on other types of sentences, including sentences with complex clausal-phrasal constructions, or sentences with non-compositional idioms, is yet to be undertaken.

\section*{Acknowledgements}
This work was partially funded by the EPSRC grant EP/T026995/1 entitled “EnnCore: End-to-End Conceptual Guarding of Neural Architectures” under Security for all in an AI enabled society, by the CRUK National Biomarker Centre, and supported by the Manchester Experimental Cancer Medicine Centre and the NIHR Manchester Biomedical Research Centre.

% The specific safety guarantees of these models are not yet fully established, and our subsequent focus will be on addressing this aspect.

% Additionally, the efficient traversal (sampling) of latent sentence spaces to exert control over generation remains a challenge, particularly given the discrete properties of sentence spaces.

% Furthermore, the unsupervised INN exhibits a distinct learning pattern for semantic distribution, a topic that warrants future exploration and explanation.

% It's important to note that this study exclusively concentrated on explanatory sentences, as presented in \cite{https://doi.org/10.48550/arxiv.2104.08661}, which effectively capture formal semantics for multi-hop natural language inference. Whether it can perform on other types of natural languages is not explored.

\bibliography{references}
\bibliographystyle{acl_natbib}

\appendix
\clearpage
\appendix

\section{Experiment setting} \label{sec:train_detail}

\paragraph{Dataset.} Table \ref{tab:stats_data} displays the statistical information of the datasets used in the experiment. The data of the two datasets partially overlap, so only the unique explanations are selected as the experimental data. 
\begin{table}[ht!]
    \small
    \centering
    \renewcommand\arraystretch{1.1}
      \resizebox{7.6cm}{!}{
    \begin{tabular}{|c|cc|}
        \hline
        Corpus & Num data. & Avg. length \\ \hline
        WorldTree \cite{jansen-etal-2018-worldtree} & 11430 & 8.65 \\
       EntailmentBank \cite{https://doi.org/10.48550/arxiv.2104.08661} & 5134 & 10.35 \\ \hline
    \end{tabular}
    }
    \caption{Statistics from explanations datasets.} \label{tab:stats_data}
\end{table}
Table \ref{tab:visua_details} illustrates the semantic, structure, and topic information of explanatory sentences over the latent space. %\cite{zhang2022}. 

The rationale for choosing explanatory sentences is that they are designed for formal/localised/symbolic semantic inference task in natural language form, %\cite{zhang2023towards}, 
which provides a semantically complex and yet controlled experimental setting, containing a both well-scoped and diverse set of target concepts and sentence structures, providing a semantically challenging yet sufficiently well-scoped scenario to evaluate the syntactic and semantic organisation of the space. 
%Besides, compared with other datasets, such as Wikipedia and Wordnet, that focus on word knowledge, it has limited semantic structures, such as atomic sentences (\textit{animal is a kind of living thing}) and conditional sentences (\textit{if a habitat is removed then that habitat is destroyed}), and limited lexical information, leading to low information entropy(disorder), potentially resulting in better semantic and structure separability captured by the latent space of INN. 
More details about semantic structure and lexical information are provided in Table \ref{tab:visua_details} and \ref{tab:srl_silva}. 
% The semantic role label annotation is done via pretrained model \cite{Shi2019SimpleBM}, which can be implemented via \textit{AllenNLP} library \cite{gardner2018allennlp}\footnote{\url{https://allenai.org/allennlp}}.

\begin{table*}[h]
    \small \setlength\tabcolsep{4.5pt}
    % \scriptsize
    \centering
\renewcommand\arraystretch{1.1}
    \begin{tabular}{|p{1cm}|p{14cm}|}  \hline
        \textbf{Cluster}         & \textbf{Theme, Pattern, and Explanatory sentences}                       \\ \hline
        0 & Theme: physics and chemistry. Pattern: \textit{if then} and \textit{as}. E.g., if a substance is mixed with another substance then those substances will undergo physical change.   \\ \hline
        1 & Theme: country, astronomy, and weather. E.g., new york state is on earth \\ \hline
        2 & Theme: physics and chemistry. Pattern: \textit{is a kind of}. E.g., light is a kind of wave. \\ \hline
        3 & Theme: biology. E.g., a mother births offspring. \\ \hline
        4 & Theme: synonym for verb. Pattern: \textit{means} and \textit{is similar to}. E.g., to report means to show. \\ \hline
        5 & Theme: astronomy. E.g., the solar system contains asteroids.\\ \hline
        6 & Theme: animal/plant. Pattern: \textit{is a kind of}. E.g., a seed is a part of a plant. \\ \hline
        7 & Theme: item. E.g., a telephone is a kind of electrical device for communication.\\ \hline
        8 & Theme: synonym for life. Pattern: \textit{means} and \textit{is similar to}. E.g., shape is a kind of characteristic.  \\ \hline
        9 & Theme: geography. Pattern: \textit{is a kind of}. E.g., a mountain is a kind of environment.\\ \hline
        10 & Theme: animal and plant. Pattern: \textit{if then} and \textit{as}. E.g., if a habitat is removed then that habitat is destroyed.  \\ \hline
        11 & Theme: scientific knowledge. Pattern: \textit{(;)}, \textit{number} and \textit{/}. E.g., freezing point is a property of a ( substance ; material ). \\ \hline
        12 & Theme: item. Pattern: \textit{is a kind of object}. E.g., a paper is a kind of object. \\ \hline
        13 & Theme: chemistry and astronomy. E.g., oxygen gas is made of only oxygen element.\\ \hline
        14 & Theme: general about science. Pattern: \textit{(;)}. E.g., seed dispersal has a positive impact on ( a plant ; a plant 's reproduction). \\ \hline
        15 & Theme: item. Pattern: \textit{is a kind of}. E.g., fertilizer is a kind of substance. \\ \hline
        16 & Theme: physics and chemistry. Pattern: \textit{(;)}. E.g., the melting point of oxygen is -3618f ; -2188c ; 544k. \\ \hline
        17 & Theme: animal. E.g., squirrels live in forests. \\ \hline
        18 & Theme: nature. E.g., warm ocean currents move to cooler ocean regions by convection.\\ \hline
        19 & Theme: life. E.g., pond water contains microscopic living organisms.\\ \hline
    \end{tabular}
    \caption{Semantic, structure, topic information of explanatory sentences, where the cluster is the categories of k-means classifier.} 
    \label{tab:visua_details}
\end{table*}
% \begin{table*}[ht!]
% \scriptsize
% \begin{center}
% \begin{tikzpicture}
% \node (table) [inner sep=.1pt] {

\begin{table*}[h]
%     \small \setlength\tabcolsep{4.5pt}
    % \scriptsize
    \small
    \centering
\renewcommand\arraystretch{1.1}
    \begin{tabular}{p{3cm}p{2cm}p{9.7cm}}  \toprule
        \textbf{Semantic Tags}     & \textbf{Prop. \%}    & \textbf{Description and Example}                       \\ \hline
        ARGM-DIR & 0.80 & Directionals. E.g. all waves transmit energy \textbf{from one place to another}  \\ \hline
        ARGM-PNC & 0.08 & Purpose. E.g. many animals blend in with their environment \textbf{to not be seen by predators} \\ \hline
        ARGM-CAU & 0.05 & Cause. E.g. cold environments sometimes are white in color \textbf{from being covered in snow} \\ \hline
        ARGM-PRP & 1.30 & Purpose. E.g. a pot is made of metal \textbf{for cooking} \\ \hline
        ARGM-EXT & 0.04 & Extent. E.g. as the amount of oxygen exposed to a fire increases the fire will burn \textbf{longer} \\ \hline
        ARGM-LOC & 4.50 & Location. E.g. a solute can be dissolved \textbf{in a solvent} when they are combined \\ \hline
        ARGM-MNR & 2.00 & Manner. E.g. fast means \textbf{quickly} \\ \hline
        ARGM-MOD & 9.80 & Modal verbs. E.g. atom \textbf{can} not be divided into smaller substances \\ \hline
        ARGM-DIS & 0.07 & Discourse. E.g. if something required by an organism is depleted \textbf{then} that organism must replenish that something \\ \hline
        ARGM-GOL & 0.20 & Goal. E.g. We flew \textbf{to Chicago} \\ \hline
        ARGM-NEG & 1.20 & Negation. E.g. cactus wrens building nests in cholla cacti does \textbf{not} harm the cholla cacti \\ \hline
        ARGM-ADV & 6.70 & Adverbials \\ \hline
        ARGM-PRD & 0.20 & Markers of secondary predication. E.g. \\ \hline
        ARGM-TMP & 7.00 & Temporals. E.g. a predator \textbf{usually} kills its prey to eat it \\ \hline
        O & - & Empty tag. \\ \hline
        V & 100 & Verb. \\ \hline
        ARG0 & 32.0 & Agent or Causer. E.g. \textbf{rabbits} eat plants \\ \hline
        ARG1 & 98.5 & Patient or Theme. E.g. rabbits eat \textbf{plants} \\ \hline
        ARG2 & 60.9 & indirect object / beneficiary / instrument / attribute / end state. E.g. animals are \textbf{organisms} \\ \hline
        ARG3 & 0.60 & start point / beneficiary / instrument / attribute. E.g. sleeping bags are designed \textbf{to keep people warm} \\ \hline
        ARG4 & 0.10 & end point. E.g. when water falls from the sky that water usually returns \textbf{to the soil} \\ \toprule
        
    \end{tabular}
    \caption{Semantic Role Labels that appear in explanations corpus.} 
    \label{tab:srl_silva}
\end{table*}

\begin{figure}[ht!]
\begin{center}
    \includegraphics[width=\columnwidth]{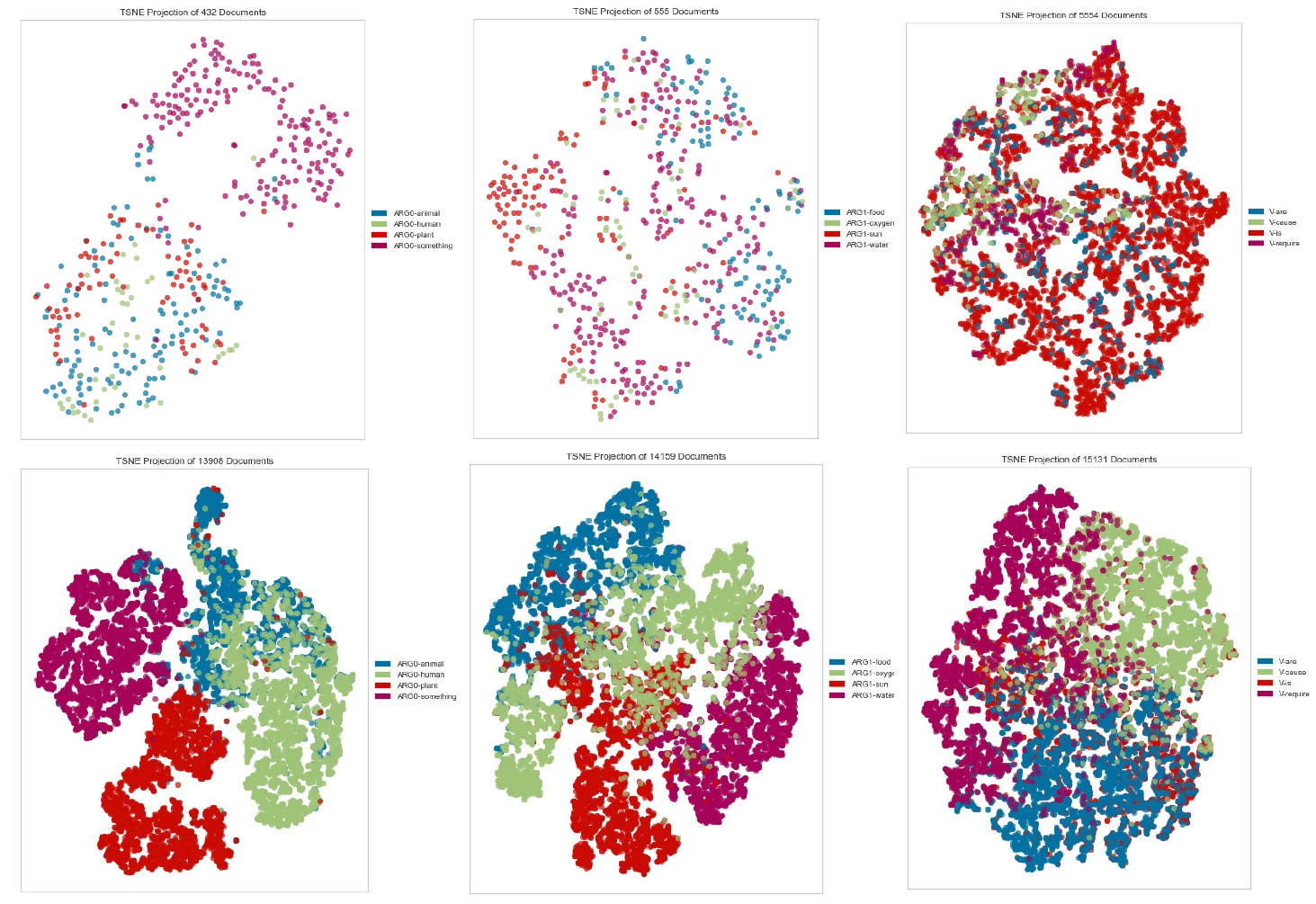}
    \caption{t-SNE plot for Data augmentation (top: original dataset distribution, bottom: augmented dataset distribution), (left: ARG0-animal(blue), human(green), plant(red), something(purple); middle: ARG1-food(blue), oxygen(green), sun(red), water(purple); right: PRED-are(blue), cause(green), is(red), require(purple).}
    \label{fig:data_augmentation}
 \end{center}
\end{figure}

\paragraph{Data Augmentation.} Algorithm \ref{alg:data_augmentation} illustrates the detailed process of data augmentation. The key aspect of data augmentation is to keep the data distribution unchanged while increasing the size of the dataset. Therefore, during traversal, we only sample the value whose probability density is between 0.495 and 0.505. In other words, for each original explanation, we only traverse its 
 close neighbours over the latent space. We increased the number of explanations in each role-content cluster to 3000 and kept the balance of each role-content category. We provide more qualitative examples in Table \ref{tab:aug_data_app}. Moreover, we visualise latent semantic distribution before and after augmentation in Figure \ref{fig:data_augmentation}. As we can observe, the data augmentation can maintain the semantic distribution unchanged. For example, \textit{PRED-is} (red colour in the right column) is widely distributed over the latent space before and after augmentation. \textit{ARG0-something} (purple colour in the left column) is far from other clusters with or without data augmentation in latent space.

\begin{algorithm}
    \caption{Data Augmentation} \label{alg:data_augmentation}
    \begin{algorithmic}
        \State \textbf{Define:} $R$ as the role set (ARG0, PRED, ...).
        \State \textbf{Define:} $C$ as the content set (vocabulary).
        \State \textbf{Define:} $S$ as the explanation corpus (sentences).
        \State \textbf{Define:} $s = [(c_1, r_1), ..., (c_i, r_i)] \in S,~ c_i \in C,~ r_i \in R$ as a sentence.
        \State \textbf{Define:} $(c_t, r_t) \quad | \quad r_t \in R,~c_t \in C$ as the target role-content (e.g., ARG1-animal).
        \State \textbf{Define:} $S_t = \forall{s \in S} \quad | \quad \exists{(c_k, r_k) = (c_t, r_t)}$ as the set of sentences with the target role-content. 
        \State \textbf{Define:} $E(s): S \rightarrow \mathbb{R}^n$ as encoder (embedding) function.
        \State \textbf{Define:} $D(vec): \mathbb{R}^n \rightarrow S$ as the explanation decoded from Decoder $D$.
        \State \textbf{Define:} $L$: list for keeping augmented sentences.
        \State \textbf{Define:} $SRLer(s)$: semantic role label annotator for $s$.
        \ForAll{$ (s_i, s_j) \in S_t,~ s_i \neq s_j$}
                \State $vec = average(E(s_i), E(s_j))$
                \ForAll{$ vec[i] \in vec$}
                    \State $vec[i] = N(0, 1)$ \# \textit{neighbour traversal}
                    \State $s_n = D(vec)$ \# \textit{new sentence}
                    \If{$s_n \notin L$ \textbf{AND} $R \in SRLer(s_n)$}
                        \State put $s_n$ in $L$.
                    \EndIf
                \EndFor
        \EndFor
    \end{algorithmic}
\end{algorithm}

% \begin{table}[t]
%     \small
%     \centering
%       \resizebox{7.6cm}{!}{
%     \begin{tabular}{|c|cc|}
%         \hline
%         Task & Num data. & Avg. length \\ \hline
%         Explanation Generation & 11430 & 8.65 \\
%       	Natural Language Inference & 5134 & - \\ \hline
%     \end{tabular}
%     }
%     \caption{Statistics from explanations datasets.} \label{tab:stats_data}
% \end{table}

\begin{table}[ht!]
    \small
    \centering
      \resizebox{7.8cm}{!}{
    \begin{tabular}{ll}
        \toprule
        Role-content & Augmented sentences \\ \hline
        \multirow{3}{*}{ARG0-plant} 
        & \textcolor{blue}{plants} use sunlight often to make food for themselves \\
        & \textcolor{blue}{plants} produce light in the winter by photosynthesizing \\
        & \textcolor{blue}{green plants} contain ( water ; food ) \\
        & \textcolor{blue}{plants} take in oxygen from the air \\
        & \textcolor{blue}{a plant} requires water in order to perform photosynthesis \\ 
        & \textcolor{blue}{some plants} grow organically \\ 
        & \textcolor{blue}{plants} use soil as a source of water \\ \hline
        \multirow{3}{*}{ARG1-water} 
        & \textcolor{blue}{water} is liquid by volume \\
        & \textcolor{blue}{salt water} is a kind of solution \\
        & \textcolor{blue}{water} is two things together  \\
        & \textcolor{blue}{water} is boiling in the pot  \\
        & \textcolor{blue}{water} is an ( inexhaustible ; wasteable ) resource \\ 
        & \textcolor{blue}{water} is an ( electrical ; electrical energy ) insulator \\
        & \textcolor{blue}{water} is a part of soup \\
        \hline
        \multirow{3}{*}{ARG2-animal} 
        & a hurricane is a kind of \textcolor{blue}{animal} \\
        & a bird is a kind of \textcolor{blue}{animal} \\
        & a sperm whale is a kind of \textcolor{blue}{animal} \\
        & a wren is a kind of \textcolor{blue}{animal} \\
        & a dog is a kind of native \textcolor{blue}{animal} \\ 
        & a chameleon is a kind of \textcolor{blue}{animal} \\ \hline
        \multirow{3}{*}{PRED-require} 
        & making tools \textcolor{blue}{requires} using sharp tools \\
        & plants \textcolor{blue}{require} resources to provide food for themselves \\
        & a system \textcolor{blue}{requires} electrical energy to operate \\
        & crops \textcolor{blue}{require} specialized environments to grow \\
        & cooking \textcolor{blue}{requires} food from human food chain \\ 
        & producing an object \textcolor{blue}{requires} chemical energy \\ 
        & living things \textcolor{blue}{require} energy from the sun for survival \\ 
        & growth \textcolor{blue}{requires} the production of more cells \\ \toprule
        
    \end{tabular}
    }
    \caption{Qualitative evaluation of geometrical data augmentation.} \label{tab:aug_data_app}
\end{table}

\paragraph{Downstream Classifier.} In this experiment, we apply three downstream classifiers, including non-parametric classifier: k-nearest neighbours (KNN) and parametric classifiers: Naive Bayes (NB) and Support Vector Machine (SVM), to evaluate the separability of latent representation. Those classifiers and classification metrics are implemented based on \textit{scikit-learn} package \cite{scikit-learn} with default hyper-parameters. We train those classifiers on the training set ($\times 60\%$) and evaluate them on the test set ($\times 40\%$). For multi-class classification, we set \textit{macro} for \textit{precision}, \textit{recall}, and \textit{f1} since \textit{macro}-averaged metric for each class is calculated independently, and then the average is taken, which ensures that the performance of the model in each class contributes equally to the final metric, regardless of the class size.

\paragraph{Visualizer.} In this experiment, we implement t-SNE and PCA visualisation based on \textit{Yellowbrick} library \cite{bengfort_yellowbrick_2019}\footnote{\url{https://www.scikit-yb.org/en/latest/api/text/tsne.html}}. We empirically set $decompose\_by=4$ for all cases. However, we found no significant difference between different $decompose\_by$ parameters.

\paragraph{Baselines for Interpolation Smoothness.} In the experiment, we implement five LSTM-based autoencoders, including denoising AE (\citet{10.1145/1390156.1390294}, DAE), $\beta$-VAE \cite{Higgins2016betaVAELB}, adversarial AE (\citet{makhzani2016adversarial}, AAE), label adversarial AE (\citet{rubenstein2018latent}, LAAE), and denoising adversarial autoencoder (\citet{shen2020educating}, DAAE). Their implementation relies on the open-source codebase available at the URL \footnote{\url{https://github.com/shentianxiao/text-autoencoders}}. As for transformer-based VAEs, we implement Optimus \cite{li2020optimus}, AdaVAE \cite{tu2022adavae}\footnote{\url{https://github.com/ImKeTT/AdaVAE}}, and Della \cite{hu-etal-2022-fuse}\footnote{\url{https://github.com/OpenVLG/DELLA}}.
% For transformer-based baselines, T5VQVAE \cite{Zhang2024ImprovingSC} is instantiated based on the official code repository, accessible through the URL \footnote{\url{https://github.com/SnowYJ/T5VQVAE}}. Della \cite{hu-etal-2022-fuse} is constructed utilizing the codebase\footnote{\url{https://github.com/OpenVLG/DELLA}}. AdaVAE \cite{tu2022adavae} is implemented based on the open code base\footnote{\url{https://github.com/ImKeTT/AdaVAE}}. 
All baseline models undergo training and evaluation with the hyper-parameters provided by their respective sources. A latent dimension of 32 is specified to ensure a uniform and equitable comparative analysis.

\paragraph{Autoencoder.} \label{app:autoencoder} In this work, we employ an autoencoder architecture with the same configuration as described in \cite{li2020optimus}\footnote{\url{https://github.com/ChunyuanLI/Optimus}}. The encoder component is based on BERT \cite{devlin2019bert}, while the decoder component is based on GPT2 \cite{Radford2019LanguageMA}. The latent space dimension is set to 32 (low-dimension) as \citet{michlo2023overlooked} revealed that strong compression, such as strong KL regularisation term in ELBO, can lead to the phenomenon of disentanglement of images.

To establish the connection between the encoder and decoder, the input sentence $x$ is first encoded by BERT[cls] into the latent space, denoted as $N(\mu, \Sigma)$. The parameters $\mu$ and $\Sigma$ are trainable and determine the mean and covariance of the Gaussian distribution. Next, a sample $z \sim N(\mu, \Sigma)$ is passed through a multi-layer perceptron called $W$. This step expands the dimensionality of $z$ to obtain a fixed-length embedding $h \in \mathbb{R}^{D \times L \times H}$, where $D$ represents the dimensions of the heads, $L$ is the number of heads, and $H$ is the number of hidden layers. The latent space injection can be described as:
\[
\begin{aligned}
    \text{Attention}(Q, K, V) = \text{softmax}( \frac{Q [z;K]^T}{\sqrt{d}})[z;V]
\end{aligned}
\]
% Figure \ref{fig:optimus} provides a visual representation of the connection between BERT and GPT2 within the AutoEncoder architecture.
% \begin{figure}[ht!]
% \begin{center}
%     \includegraphics[width=\columnwidth]{figs/optimus.png}
%     \caption{Latent sentence injection.}
%     \label{fig:optimus}
%  \end{center}
% \end{figure}

\begin{figure}[ht!]
\begin{center}
    \includegraphics[width=\columnwidth]{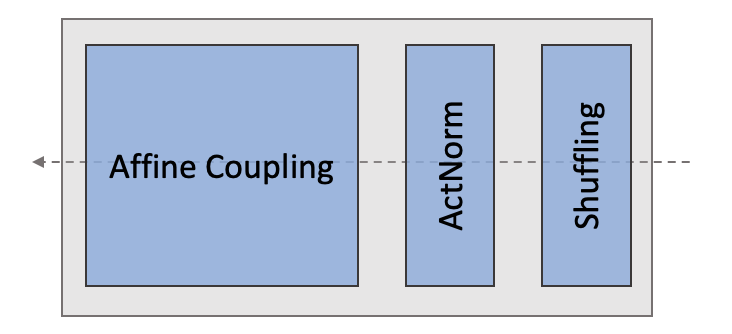}
    \caption{INN one single block.}
    \label{fig:inn_block}
 \end{center}
\end{figure}

\paragraph{INN.} The INN consists of 10 invertible blocks. Each is built from three layers, including an affine coupling \cite{dinh2016density}, permutation layer, and ActNorm \cite{kingma2018glow}. Figure \ref{fig:inn_block} displays one single invertible block. The model was implemented using the FrEIA library \cite{freia} \footnote{\url{https://github.com/VLL-HD/FrEIA}}. As for training hyperparameters of INN, firstly, both input and output have the same dimensions as the latent space dimension of the autoencoder. Secondly, inside the affine coupling block, the sub-network is MLP with 512 as the hidden dimension. Thirdly, we use AdamW \cite{https://doi.org/10.48550/arxiv.1711.05101} to optimise the model where the learning rate is 5e-04 in the experiment. 

\noindent The forward process of the affine coupling layer can be described as follows: 
\begin{equation}
\begin{split}
x_a, x_b = \text{split}(x) \\
\log s, t = m_{\theta}(x_b) \\
s = \exp (\log s) \\
y_a = s \odot x_a + t \\
y_b = x_b \\
y = \text{concat}(y_a, y_b)
\end{split}
\end{equation}
Where $m_{\theta}$ is a two-layer neural network. $x$ and $y$ are the input and output. The reversed process is:
\begin{equation}
\begin{split}
y_a, y_b = \text{split}(y) \\
\log s, t = m_{\theta}(y_b) \\
s = \exp (\log s) \\
x_a = (y_a - t) / s \\
x_b = y_b \\
y = \text{concat}(x_a, x_b)
\end{split}
\end{equation}

\section{Additional Supervision Results} \label{sec:dis_pred}
\begin{figure}[ht!]
% \begin{center}
\centering
    \includegraphics[scale=0.16]{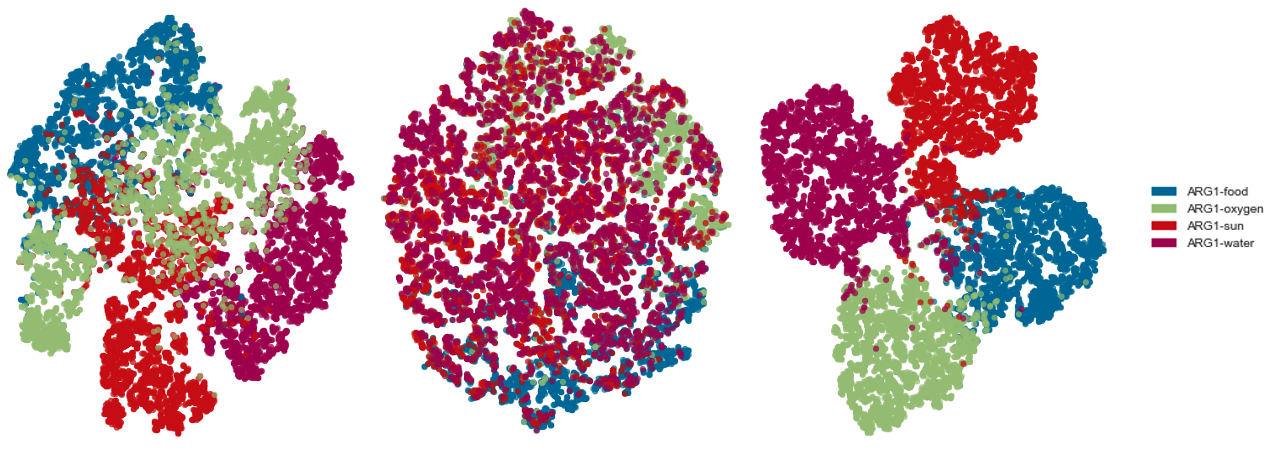}
    \caption{ARG1: t-SNE plot (blue: \textit{food}, green: \textit{oxygen}, red: \textit{sun}, purple: \textit{water}). Supervision (right) induces separability comparable with ARG0. PCA plot is provided in Figure \ref{fig:a1_pca}.}
    \label{fig:food_water}
 % \end{center}
\end{figure}

\paragraph{Disentanglement between \textit{ARG1} clusters} We consider four \textit{ARG1} clusters, including \textit{ARG1-food}, \textit{ARG1-oxygen}, \textit{ARG1-sun}, \textit{ARG1-water}, and evaluate model performance following the same procedure. Figure \ref{fig:food_water} displays the distributions of four role-content clusters over the latent space. With similar observations as before, the INN cluster-supervised training strategy can learn better disentanglement between ARG1 clusters. 
% Additionally, when compared with the ARG0 cluster, the Optimus model does not show observable disentanglement.
Table \ref{tab:arg1_exp} and \ref{tab:arg1_invert_ratio} show the disentanglement metrics and invertibility ratio, respectively. With similar observations as the previous experiment: all classifiers trained over the supervised latent representation outperform both the unsupervised INN model and Optimus, and both unsupervised and supervised cases can achieve higher ratios (at least 0.95).
% \begin{table}[ht!]
% % \scriptsize
% \small
% \centering
% \setlength\tabcolsep{2.5pt}
% \resizebox{7.8cm}{!}{
% \renewcommand\arraystretch{1}
% \begin{tabular}{cccccc}
% \toprule
% \multicolumn{6}{c}{ARG1: disentanglement proxy metrics (forward: $T$)} \\ \hline

% classifier & train & accuracy & precision & recall  & f1 score \\ \hline
% \multirow{3}{*}{KNN}  & O & 0.958 & 0.958 & 0.958  & 0.958 \\
% & U & 0.951 & 0.951 & 0.951  & 0.951 \\
% & C & \textbf{0.969} & \textbf{0.969} & \textbf{0.969}  & \textbf{0.969}  \\ \hline
 
% \multirow{3}{*}{NB} & O & 0.907 & 0.907 & 0.907  & 0.907 \\
% & U & 0.926 & 0.926 & 0.926  & 0.926 \\
% & C & \textbf{0.956} & \textbf{0.956} & \textbf{0.956}  & \textbf{0.956} \\ \hline
 
% \multirow{3}{*}{SVM} & O & 0.956 & 0.956 & 0.956  & 0.956 \\
% & U & 0.953 & 0.953 & 0.953 & 0.953 \\
% & C & \textbf{0.958} & \textbf{0.958} & \textbf{0.958}  & \textbf{0.958} \\ \toprule

% % \multicolumn{6}{c}{ARG1: invertibility ratio (backward: $T'$)} \\ \hline
% % \multicolumn{2}{c}{train} & food & oxygen & sun & water \\ \hline
% % \multicolumn{2}{c}{U} & \textbf{0.990} & \textbf{0.980} & 0.950 & 1.000 \\ 
% % \multicolumn{2}{c}{C} & 0.960 & 0.950 & \textbf{0.960} & \textbf{1.000} \\ \toprule
% \end{tabular}
% }
% \caption{Forward evaluation for ARG1, consistent results on different classifiers indicate that supervision can perform better semantic disentanglement.} \label{tab:arg1_exp}
% \end{table}

\begin{table}[ht!]
% \scriptsize
\small
\centering
\setlength\tabcolsep{2.5pt}
\resizebox{7.8cm}{!}{
\renewcommand\arraystretch{1}
\begin{tabular}{cccccc}
\toprule
\multicolumn{6}{c}{ARG1: disentanglement proxy metrics (forward: $T$)} \\ \hline

classifier & train & accuracy & precision & recall  & f1 score \\ \hline
\multirow{3}{*}{KNN}  & O & 0.934 & 0.934 & 0.933  & 0.933 \\
& U & 0.914 & 0.914 & 0.914  & 0.913 \\
& C & \textbf{0.954} & \textbf{0.954} & \textbf{0.954}  & \textbf{0.954}  \\ \hline
 
\multirow{3}{*}{NB} & O & 0.904 & 0.910 & 0.902  & 0.904 \\
& U & 0.922 & 0.922 & 0.922  & 0.922 \\
& C & \textbf{0.957} & \textbf{0.957} & \textbf{0.957}  & \textbf{0.957} \\ \hline
 
\multirow{3}{*}{SVM} & O & 0.951 & 0.951 & 0.951  & 0.950 \\
& U & 0.953 & 0.953 & 0.952 & 0.953 \\
& C & \textbf{0.959} & \textbf{0.959} & \textbf{0.959}  & \textbf{0.959} \\ \toprule

% \multicolumn{6}{c}{ARG1: invertibility ratio (backward: $T'$)} \\ \hline
% \multicolumn{2}{c}{train} & food & oxygen & sun & water \\ \hline
% \multicolumn{2}{c}{U} & \textbf{0.990} & \textbf{0.980} & 0.950 & 1.000 \\ 
% \multicolumn{2}{c}{C} & 0.960 & 0.950 & \textbf{0.960} & \textbf{1.000} \\ \toprule
\end{tabular}
}
\caption{Forward evaluation for ARG1, consistent results on different classifiers indicate that supervision can perform better semantic disentanglement.} \label{tab:arg1_exp}
\end{table}

% Table \ref{tab:arg1_exp} shows the disentanglement metrics (top) and invertibility ratio (bottom). With similar observations as the previous experiment: all classifiers trained over the supervised latent representation outperform both the unsupervised INN model and Optimus, and both unsupervised and supervised cases can achieve higher ratios (at least 0.95).
% \input{tables/arg1_all.tex}
% \multicolumn{6}{c}{ARG1: invertibility ratio (backward: $T'$)} \\ \hline
% \multicolumn{2}{c}{train} & food & oxygen & sun & water \\ \hline
% \multicolumn{2}{c}{U} & \textbf{0.990} & \textbf{0.980} & 0.950 & 1.000 \\ 
% \multicolumn{2}{c}{C} & 0.960 & 0.950 & \textbf{0.960} & \textbf{1.000} \\
\paragraph{Invertibility ratio.} Table \ref{tab:arg1_invert_ratio}, \ref{tab:v_invert_ratio}, and \ref{tab:animal_invert_ratio} report the invertibility test for \textit{ARG1}, \textit{PRED}, and \textit{ARG0,1,2} clusters, respectively. We can observe that INN with both training approaches can perform stable invertibility. 

\begin{table}[ht]
\small
\centering
\setlength\tabcolsep{2.5pt}
\begin{tabular}{cccccc}
\toprule 
\multicolumn{6}{c}{ARG1: invertibility ratio (backward: $T'$)} \\ \hline
\multicolumn{2}{c}{train} & food & oxygen & sun & water \\ \hline
\multicolumn{2}{c}{U} & 0.990 & 0.980 & 0.950 & 1.000 \\ 
\multicolumn{2}{c}{C} & 0.960 & 0.950 & 0.960 & 1.000 \\\toprule
\end{tabular}
\caption{backward evaluation for ARG1 clusters. unsupervised INN (U), and supervised INN (S).} \label{tab:arg1_invert_ratio}
\end{table}
\begin{table}[ht]
\small
\centering
\setlength\tabcolsep{2.5pt}
\begin{tabular}{cccccc}
\toprule 
 \multicolumn{6}{c}{PRED: invertibility test (backward: $T'$)} \\ \hline
\multicolumn{2}{c}{train} & is & are & cause & require \\ \hline
\multicolumn{2}{c}{U} & 1.000 & 0.950 & 0.970 & 0.800 \\
\multicolumn{2}{c}{C} & 1.000 & 0.880 & 0.900 & 0.820 \\ \toprule
\end{tabular}
\caption{backward evaluation for predicate clusters. unsupervised INN (U), and supervised INN (S).} \label{tab:v_invert_ratio}
\end{table}

\begin{table}[ht!]
\small
\centering
\setlength\tabcolsep{2.5pt}
\renewcommand\arraystretch{1}
\begin{tabular}{cccc}
\toprule
\multicolumn{4}{c}{Animal: invertibility ratio (backward: $T'$)} \\ \hline
\multicolumn{1}{c}{train} & ARG0 & ARG1 & ARG2 \\ \hline
\multicolumn{1}{c}{U} & 0.990 & 0.990 & 0.900 \\
\multicolumn{1}{c}{C} & 0.970 & 0.960 & 0.920 \\ \toprule
\end{tabular}
\caption{Backward evaluation for Animal.} \label{tab:animal_invert_ratio}
\end{table}

% We analyse the disentanglement between PREDICATE (PRED) clusters. Figure \ref{fig:v_sup} shows the distribution of four PRED clusters, including \textit{is}, \textit{are}, \textit{cause}, and \textit{require}, over latent space. Although the disentanglement of PRED clusters is not as high as ARG0 or ARG1, the latent space with cluster supervision still performs better than both the unsupervised case and the Optimus model. Table \ref{tab:v_exp}, the supervised INN model achieves better disentanglement and both unsupervised and supervised could obtain a higher ratio.
% \begin{figure}[ht!]
% \begin{center}
%     \includegraphics[scale=0.15]{figs/v_1.png}
%     \caption{PRED: t-SNE plot (blue: are, green: cause, red: is, purple: require) (left: Optimus, middle: unsupervised, right: cluster supervised).}
%     \label{fig:v_sup}
%  \end{center}
% \end{figure}
% \input{tables/v_all}

% Table \ref{tab:middle_examples} provides the middle points between \textit{ARG0-human} and \textit{ARG0-animal} clusters.
% \input{tables/middle_explainations.tex}

\paragraph{Traversal decoding for \textit{Animal} clusters.} Table \ref{tab:animal_example} shows the decoded explanations traversed around the central point of each cluster in the latent space of cluster-supervised INN. 
\begin{table}[ht!]
\begin{tcolorbox}[fontupper=\small, fontlower=\small, middle=0.3cm, title=Traversing Animal clusters]
1: \textcolor{blue}{animals} must escape from predators \\ 
2: \textcolor{blue}{animals} require air to breathe \\
3: \textcolor{blue}{an animal} requires warmth for survival \\

% 4: \textcolor{blue}{animals} that eat other animals will not have the same ability to digest bacteria \\ and feed on fresh food
% 5. \textcolor{blue}{most animals} eat food that has a direct impact on a living thing \\
1: \textcolor{blue}{animals} are small in size \\
2: \textcolor{blue}{animals} usually are not carnivores \\
% 3: \textcolor{blue}{animals} are major animals \\
% 4: \textcolor{blue}{animals} living in the environment are a kind of organism \\
3: \textcolor{blue}{animals} are a part of an environment \\

1: a rabbit is a kind of \textcolor{blue}{animal} \\
2: an otter is a kind of \textcolor{blue}{animal} \\
3: a horse is a kind of \textcolor{blue}{animal}
% 4: a frog is a kind of organism \\
% 5: a butterfly is a kind of \textcolor{blue}{animal}
\end{tcolorbox}
\caption{Traversal in each cluster (top: \textit{ARG0-Animal}, middle: \textit{ARG1-Animal}, bottom: \textit{ARG2-Animal}).}
\label{tab:animal_example}
\end{table}

%\end{document}

\paragraph{Traversal decoding for cluster connection.} Table \ref{tab:middle_examples} displays the decoded middle points between clusters. It is also observable that there are low-density embedding regions at the transition (connection) between two clusters. We decode the middle datapoints between \textit{animal} and \textit{human} clusters and list them in Table \ref{tab:middle_examples}. From those examples, we can observe that such explanations are related to both \textit{animal} and \textit{human}. This result implies that the explanations may be geometrically represented in a similar way as they were originally designed in the WorldTree corpus (maximising lexical overlaps for pred-arg alignments within an explanation chain) for supporting multi-hop inference tasks.
%\documentclass[border=.1mm,tikz,preview]{article}
%
%\usepackage{tikz}
%\usepackage{xcolor}
%\usepackage{graphicx}
%\usepackage{multirow}
%\usetikzlibrary{positioning}
%
%\begin{document}
% \usepackage[dvipsnames]{xcolor}

\begin{table}[ht!]
% \scriptsize
% \begin{center}
% \begin{tikzpicture}
% \node (table) [inner sep=.1pt] {
% \renewcommand\arraystretch{20}
% \begin{tabular}{p{0.3cm} p{6.3cm}}
% \setlength\tabcolsep{.5pt}
% \multirow{12}{*}{\rotatebox[origin=c]{90}{DSR Optimus}}
% & \textcolor{blue}{protons are found in the nucleus of an atom} \\
% & 1 protons are found in the nucleus of an atom \\
% & 2 1 atom is positive 1 in electric charge \\
% & 3 1 in 6000 is equal to 27 in 10 years \\
% & 4 if protons and neutrons have the same number of neutrons then those two particles are physically closer than one another \\
% & 5 if a neutron has a negative -10 electric charge then the atom will not be able to move \\
% & 6 if a neutron has a negative -10 electric charge then the neutron will not have a positive electric charge \\
% & \textcolor{blue}{if a neutral atom loses an electron then an atom with a positive charge will be formed} \\

% \end{tabular}
% };
% \draw [rounded corners=.5em, line width=.5pt] (table.north west) rectangle (table.south east);
% \end{tikzpicture}
% \end{center}
\begin{tcolorbox}[fontupper=\small, fontlower=\small, title=Cluster connection]
1. \textcolor{teal}{humans} sometimes hunt \textcolor{cyan}{animals} that are covered in fur \\
% 2. a \textcolor{teal}{human} / \textcolor{cyan}{animal} requires warmth for survival \\
2. \textcolor{cyan}{animals} / \textcolor{teal}{human} habitats require food \\
3. an \textcolor{cyan}{animal} may be bred with a \textcolor{teal}{human} for food \\
4. \textcolor{cyan}{animals} eat \textcolor{teal}{humans} \\
5. a \textcolor{teal}{human} can not eat algae and other \textcolor{cyan}{animals} 
\end{tcolorbox}
\caption{Middle explanations between \textit{ARG0-animal} and \textit{ARG0-human}.}
\label{tab:middle_examples}
\end{table}

%\end{document}

\begin{figure}[ht!]
    \centering
    \includegraphics[width=\columnwidth]{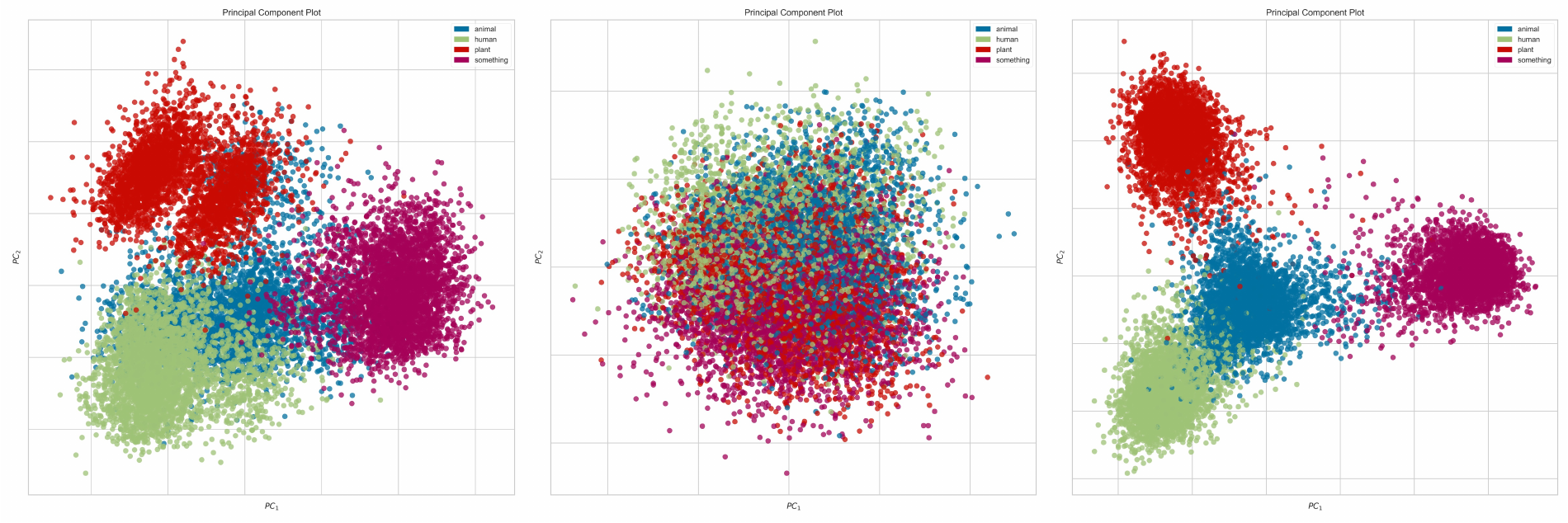}
    \caption{PCA visualization for \textit{ARG0}.}
    \label{fig:a0_pca}
\end{figure}

\begin{figure}[ht!]
    \centering
    \includegraphics[width=\columnwidth]{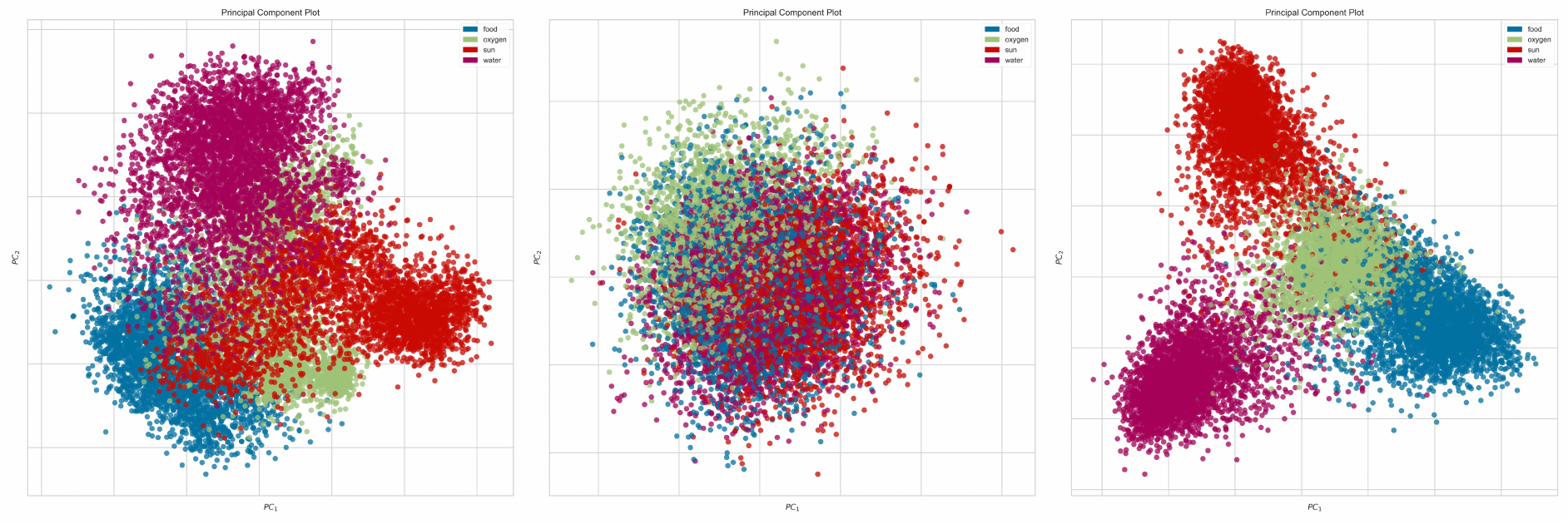}
    \caption{PCA visualization for \textit{ARG1}.}
    \label{fig:a1_pca}
\end{figure}

\begin{figure}[ht!]
    \centering
    \includegraphics[width=\columnwidth]{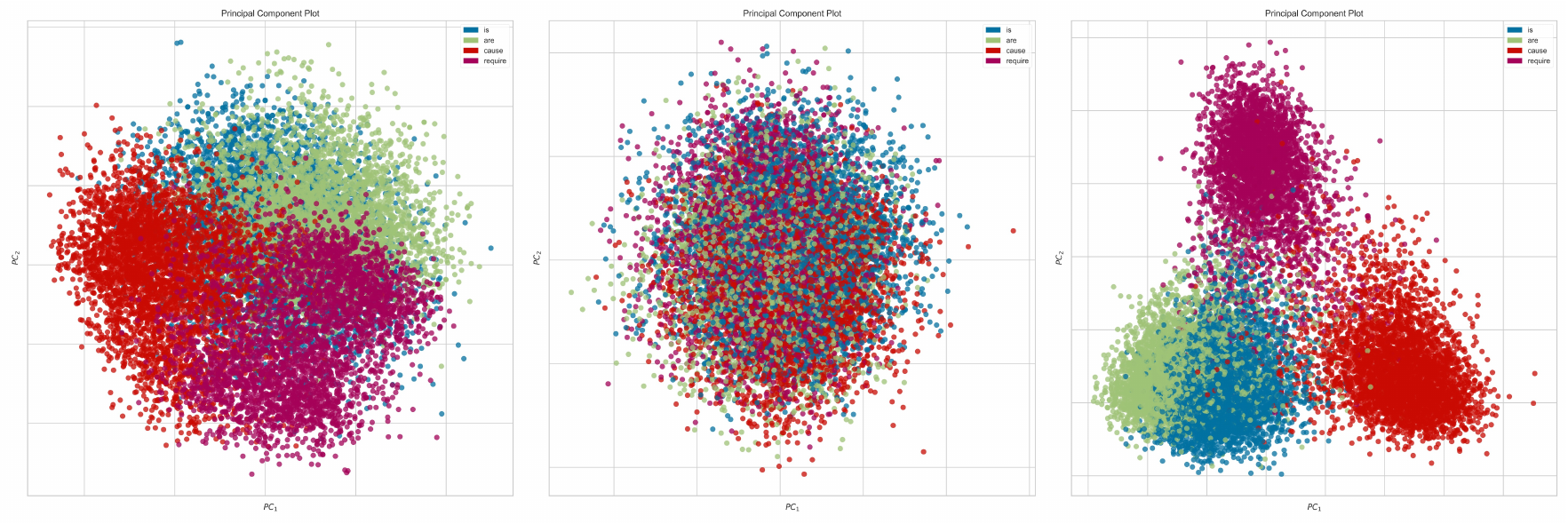}
    \caption{PCA visualization for \textit{PRED}.}
    \label{fig:verb_pca}
\end{figure}

\begin{figure}[ht!]
    \centering
    \includegraphics[width=\columnwidth]{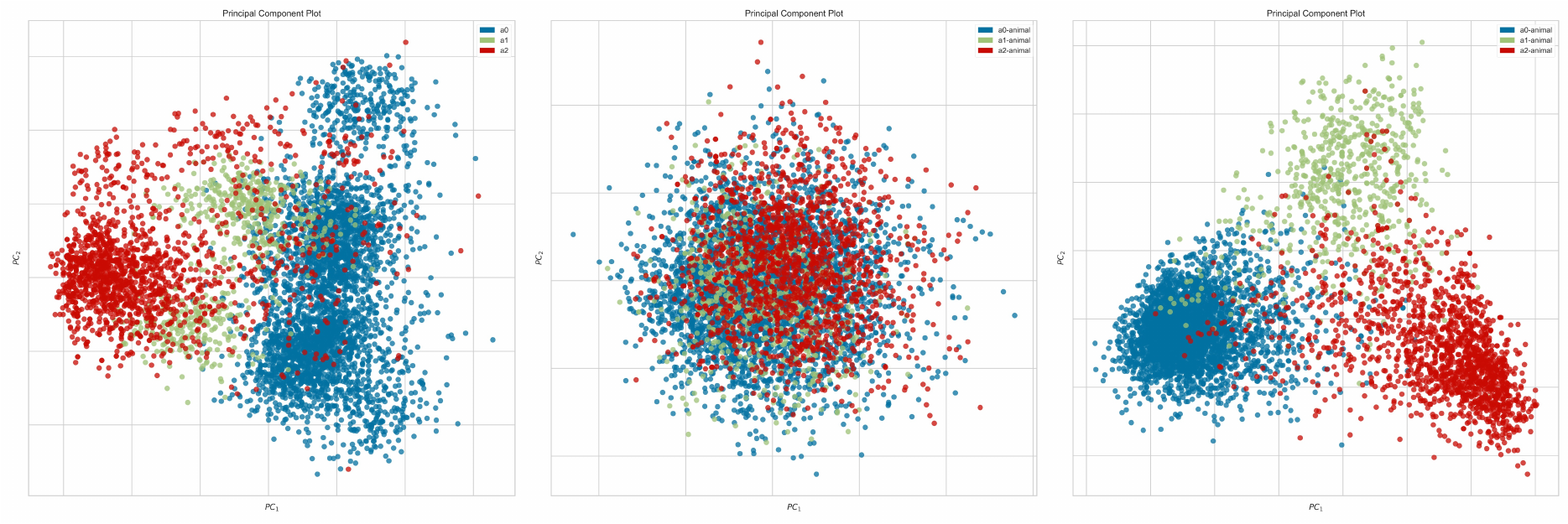}
    \caption{PCA visualization for \textit{Animal}.}
    \label{fig:animal_pca}
\end{figure}

\paragraph{Principal component analysis (PCA) visualisation.} In addition to the non-linearised t-SNE plot, we also provide linearised visualisation via PCA \cite{shlens2014tutorial}. Figure \ref{fig:a0_pca},\ref{fig:a1_pca},\ref{fig:verb_pca}, and \ref{fig:animal_pca} visualize the separation of \textit{ARG0}, \textit{ARG1}, \textit{PRED}, and \textit{animal}. Similar to the observation before, cluster supervision can lead to better separation and cluster.

\section{Statistical Significance Tests for \textit{PRED} Downstream Classifiers} \label{sec:significance}
Statistical significance testing is a standard statistical tool devised to ensure that experimental results are not coincidental and reliable. Following the work \cite{P18-1128}\footnote{\url{https://github.com/rtmdrr/testSignificanceNLP/tree/master}}, we provide statistical significance tests to rigorous and quantitatively evaluate the stability of trainable downstream classifiers, which indirectly indicates the representation capability. 

Our attention was directed towards \textit{PRED} clusters due to the comparatively decreased performance of downstream classifiers within this category as \textit{PRED} usually contains less semantic information \cite{zhang2022}. We select \textit{accuracy} metric, set $\alpha=0.05$, and choose \textit{bootstrap} statistical test which was used with a variety of NLP tasks \cite{ouchi-etal-2017-neural,wu-etal-2017-sequence}. 

As illustrated in Table \ref{tab:significant_test}, (1) the U-C pair consistently yields a diminished significance value, suggesting reliable classification performance resulting from superior representational capabilities facilitated by the AutoEncoder with INN configuration, compared with Optimus. (2) the scores of (O-C) pairs are consistently lower than those of (O-U) pairs, indicating our supervision (C) can better represent semantic information than unsupervised INN. We refer \cite{P18-1128} for an in-depth illustration of statistical significance tests in NLP.
\begin{table}[ht]
\small
\centering
\setlength\tabcolsep{2.5pt}
% \resizebox{7.8cm}{!}{
\begin{tabular}{ccc}
\toprule
\multicolumn{3}{c}{Statistical significance tests for \textit{PRED}} \\ \hline
classifier & source & Bootstrap ($p$-value)$\downarrow$ \\ \hline
\multirow{3}{*}{KNN} & O-C & 0.0155 \\
 & U-C & \textbf{\underline{0.0000}} \\
 & O-U & 1.0000\\ \hline
\multirow{3}{*}{NB} & O-C & 0.0000 \\
& U-C & \textbf{\underline{0.0000}} \\
 & O-U & 0.2268 \\ \hline
\multirow{3}{*}{SVM} & O-C & 0.3594 \\
 & U-C & \textbf{\underline{0.0000}} \\
 & O-U & 1.0000 \\ \toprule
\end{tabular}
% }
\caption{Statistical significance tests for downstream classifiers (O: Optimus, U: unsupervised INN, and C: cluster supervised INN). We \textbf{\underline{highlight}} the best significant test value, indicating reliable classification performance derived from better representation capability.} \label{tab:significant_test}
\end{table}
% \begin{table}[ht]
% \small
% \centering
% \setlength\tabcolsep{2.5pt}
% % \resizebox{7.8cm}{!}{
% \begin{tabular}{cccccc}
% \toprule
% \multicolumn{6}{c}{Statistical significance tests: Bootstrap $\downarrow$} \\ \hline
% classifier & source & \textit{ARG0} & \textit{PRED} & \textit{ARG1} & \textit{Animal} \\ \hline
% \multirow{3}{*}{KNN} & O-C & 0.0034 & 0.0155 & 0.0000 \\
%  & U-C & 0.0000 & 0.0000 & 0.0000 \\
%  & O-U & 1.0000 & 1.0000 & 1.0000 \\ \hline
% \multirow{3}{*}{NB} & O-C & 0.0000 & 0.0000 & 0.0000 \\
% & U-C & 0.0000 & 0.0000 & 0.0000 \\
%  & O-U & 0.0000 & 0.2268 & 0.0000 \\ \hline
% \multirow{3}{*}{SVM} & O-C & 0.0181 & 0.3594 & 0.0001 \\
%  & U-C & 8.9863e-06 & 0.0000 \\
%  & O-U & 0.9521 & 1.0000 \\ \toprule
% \end{tabular}
% % }
% \caption{Statistical significance tests for downstream classifiers. We \textbf{\underline{highlight}} the best significant test value, indicating reliable classification performance derived from better representation capability.} \label{tab:significant_test}
% \end{table}

\section{Ablation of Data Augmentation} \label{sec:ablation}
\begin{figure}[ht!]
    \centering
    \includegraphics[width=\columnwidth]{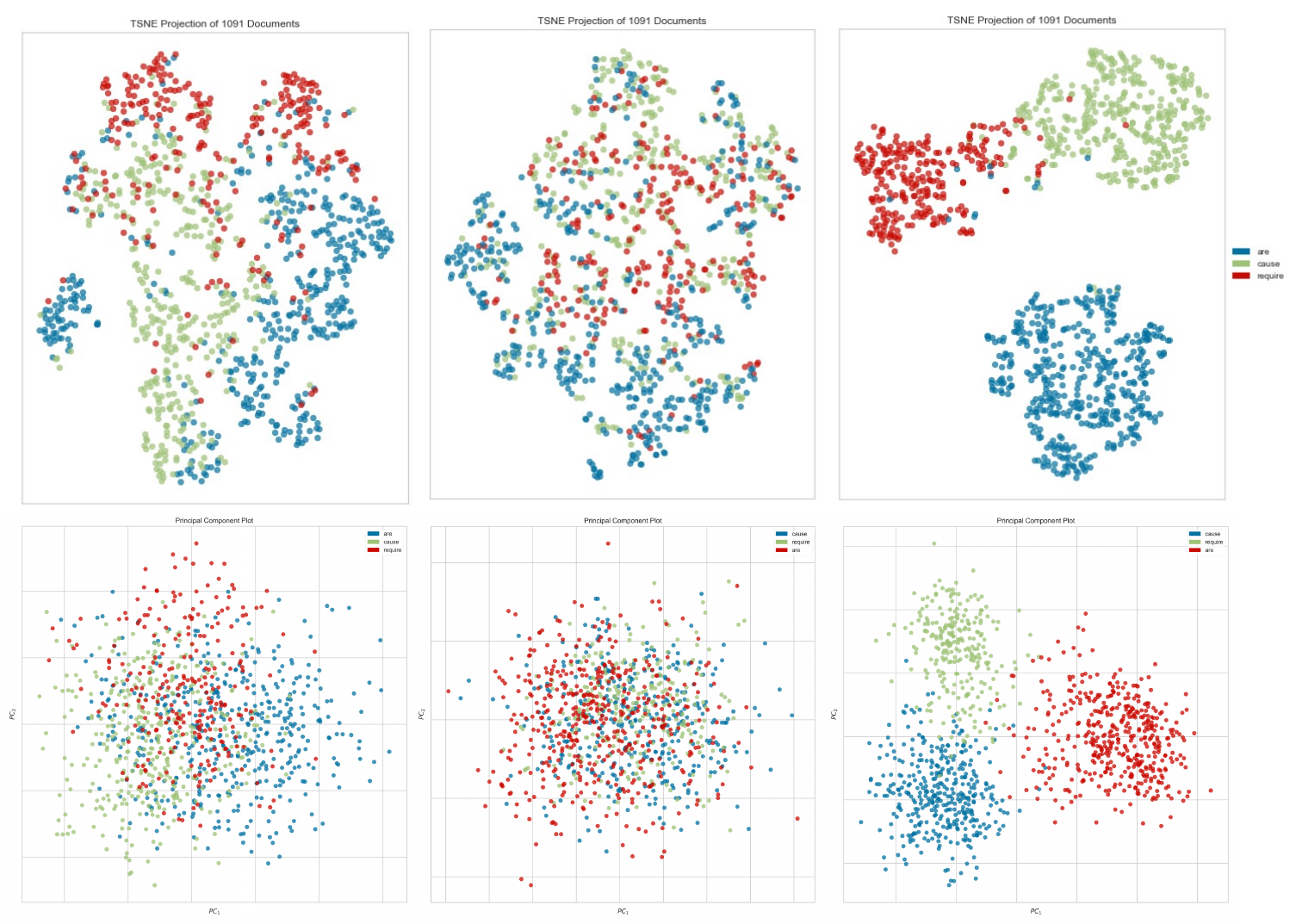}
    \caption{Ablation: t-SNE plot (top), PCA plot (bottom) (left: Optimus, middle: unsupervised, right: cluster-supervised) where blue: \textit{PRED-are}, green: \textit{PRED-cause}, red: \textit{PRED-require}.}
    \label{fig:ablation}
\end{figure}

\paragraph{\textit{PRED} semantic role.} Firstly, we analyse the effect of our supervision approach on \textit{PRED} semantic role with three lexical contents without data augmentation, including \textit{are} ($\times449$), \textit{cause} ($\times380$), and \textit{require} ($\times262$). The rationale for their selection is that they are less frequent in corpus and partially overlap in latent space. Moreover, the contents under \textit{PRED} usually have less effect on the contextual semantics \cite{zhang2022}. Those difficulties allow us to fairly analyse the effect of our supervision approach. Following a similar order, we first visualise the t-SNE and PCA plots in Figure \ref{fig:ablation}. As we can observe, the cluster-supervised approach can better represent the cluster and separation for different contents under \textit{PRED} semantic role label without data augmentation. Next, we apply downstream classifiers to evaluate cluster separation. As illustrated in Table \ref{tab:ablation_metrics}, our cluster-supervised approach results in better classification performance, indicating better disentanglement. 
% Besides, we can observe that latent representations from Optimus without data augmentation achieve lower disentanglement scores than unsupervised INN, which contradicts the aforementioned experiment that Optimus with data augmentation can lead to better classification. This result indicates that without data augmentation, AutoEncoder with unsupervised INN has the potential to better represent disentanglement than Optimus.

\begin{table}[ht]
\small
\centering
\setlength\tabcolsep{2.5pt}
\resizebox{7.8cm}{!}{
\begin{tabular}{cccccc}
\toprule
\multicolumn{6}{c}{\textit{PRED}: disentanglement proxy metrics} \\ \hline

classifier & train & accuracy & precision & recall  & f1 score \\ \hline
\multirow{3}{*}{KNN} & O & 0.858 & 0.847 & 0.844  & 0.846 \\
 & U & 0.837 & 0.849 & 0.827  & 0.830 \\
 & C & \textbf{0.965} & \textbf{0.963} & \textbf{0.961}  & \textbf{0.962} \\ \hline
 
\multirow{3}{*}{NB} & O & 0.839 & 0.823 & 0.833  & 0.826 \\
& U & 0.901 & 0.895 & 0.891  & 0.893 \\
 & C & \textbf{0.977} & \textbf{0.974} & \textbf{0.975}  & \textbf{0.974} \\ \hline
 
\multirow{3}{*}{SVM} & O & 0.876 & 0.863 & 0.866  & 0.865 \\
 & U & 0.954 & 0.953 & 0.949  & 0.950 \\
 & C & \textbf{0.967} & \textbf{0.965} & \textbf{0.967}  & \textbf{0.966} \\ \toprule
\end{tabular}
}
\caption{Ablation: disentanglement proxy metrics for \textit{PRED-are}, \textit{PRED-cause}, and \textit{PRED-require}.} \label{tab:ablation_metrics}
\end{table}

\begin{figure}[ht!]
    \centering
    \includegraphics[width=\columnwidth]{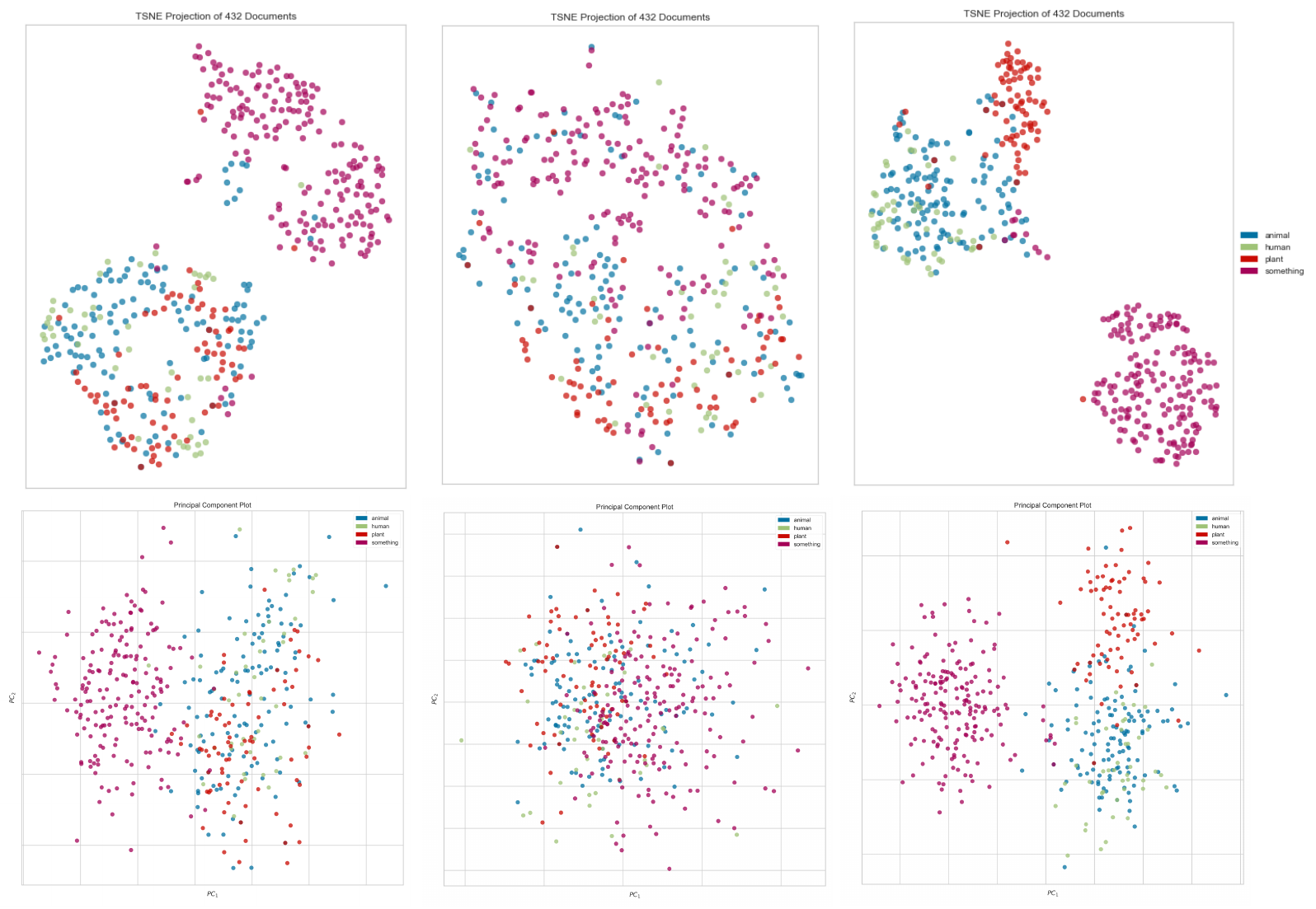}
    \caption{Ablation: t-SNE plot (top), PCA plot (bottom) (left: Optimus, middle: unsupervised, right: cluster-supervised) where blue: \textit{ARG0-animal}, green: \textit{ARG0-human}, red: \textit{ARG0-plant}, purple: \textit{ARG0-something}.}
    \label{fig:ablation_arg0}
\end{figure}

\paragraph{\textit{ARG0} semantic role.} Next, we provide the same analyse for fewer frequent \textit{ARG0} clusters: \textit{ARG0-animal} ($\times126$), \textit{ARG0-human} ($\times43$), \textit{ARG0-plant} ($\times77$), and \textit{ARG0-something} ($\times186$). As illustrated in Figure \ref{fig:ablation_arg0}, cluster supervision can lead to better role-content separation/disentanglement. Moreover, we can observe that cluster-supervision leads to better proxy disentanglement metrics in Table \ref{tab:ablation_metrics_arg0}. 

Furthermore, compared with Table \ref{tab:arg0_exp}, the incorporation of latent representation with data augmentation results in enhanced classification performance. This observation implies that our data augmentation technique can more effectively capture semantic information, thereby aiding downstream classifiers.
\begin{table}[ht]
\small
\centering
\setlength\tabcolsep{2.5pt}
\resizebox{7.8cm}{!}{
\begin{tabular}{cccccc}
\toprule
\multicolumn{6}{c}{\textit{ARG0}: disentanglement proxy metrics} \\ \hline

classifier & train & accuracy & precision & recall  & f1 score \\ \hline
\multirow{3}{*}{KNN} & O & 0.890 & 0.890 & 0.850  & 0.867 \\
 & U & 0.890 & 0.896 & 0.834  & 0.858 \\
 & C & \textbf{0.919} & \textbf{0.907} & \textbf{0.858} & \textbf{0.877} \\ \hline
 
\multirow{3}{*}{NB} & O & 0.855 & 0.809 & 0.784  & 0.792 \\
& U & 0.936 & 0.916 & 0.905 & 0.910 \\
 & C & \textbf{0.965} & \textbf{0.958} & \textbf{0.950}  & \textbf{0.954} \\ \hline
 
\multirow{3}{*}{SVM} & O & 0.843 & 0.630 & 0.691  & 0.656 \\
 & U & 0.895 & 0.847 & 0.770  & 0.782 \\
 & C & \textbf{0.901} & \textbf{0.935} & \textbf{0.779}  & \textbf{0.790} \\ \toprule
\end{tabular}
}
\caption{Ablation: disentanglement proxy metrics for \textit{ARG0-animal}, \textit{ARG0-human}, \textit{ARG0-plant}, and \textit{ARG0-something}.} \label{tab:ablation_metrics_arg0}
\end{table}

\section{Controlled Interpolation}
In tables \ref{tab:guide_generation_app2} and \ref{tab:guide_generation_app1}, we provide more controllable interpolation examples. Those examples reveal that the latent space with better role-content separation from supervised INN can provide better interpolation control, indicating better latent space geometry.
\begin{table}[ht!]
\begin{tcolorbox}[fontupper=\small, fontlower=\small, title=Interpolation localisation: \textit{predicate-require} ]
\underline{source: humans \textcolor{blue}{require} freshwater for survival}\\ \\
AutoEncoder: \\
1. humans \textcolor{blue}{require} water to survive \\
2. marine mammals \textcolor{blue}{require} great amounts of water \\
3. animals \textcolor{blue}{require} oxygen to survive  \\
4. animals \textcolor{blue}{require} water for survival \\
5. animals \textcolor{red}{must eat} water to survive \\
6. animals \textcolor{blue}{require} water and food  \\
7. animals \textcolor{blue}{require} water for survival \\
8. animals \textcolor{red}{must eat} to survive  \\
9. animals \textcolor{blue}{require} food for survival  \\
10. animals \textcolor{red}{must eat} food to survive \\

Unsupervised INN: \\
1. nonhumans \textcolor{blue}{require} water to survive \\
2. marine animals \textcolor{blue}{require} food for survival  \\
3. animals \textcolor{red}{must breath} to survive  \\
4. animals \textcolor{blue}{require} water for survival \\
5. animals \textcolor{blue}{require} water from their ecosystems  \\
6. animals \textcolor{blue}{require} water for survival \\
7. animals \textcolor{red}{must eat} food for survival \\
8. animals \textcolor{blue}{require} food for survival \\
9. animals \textcolor{blue}{require} food for survival \\
10. animals \textcolor{blue}{require} food for survival \\

% % unsupervised INN
% 1. nonhumans \textcolor{blue}{require} water to survive \\
% 2. marine animals \textcolor{blue}{require} food for survival  \\
% 3. animals \textcolor{red}{must breath} to survive  \\
% 4. animals \textcolor{blue}{require} water for survival \\
% 5. animals \textcolor{blue}{require} water from their ecosystems  \\
% 6. animals \textcolor{blue}{require} water for survival \\
% 7.: animals \textcolor{red}{must eat} food for survival \\
% 8. animals \textcolor{blue}{require} food for survival \\
% 9. animals \textcolor{blue}{require} food for survival \\
% 10. animals \textcolor{blue}{require} food for survival \\

\underline{target: animals \textcolor{blue}{require} food to survive}
\end{tcolorbox}

\caption{Interpolation examples where top and bottom sentences are source and target, respectively.}
\label{tab:guide_generation_app}
 \end{table}

\begin{table*}[ht!]
\centering
% \resizebox{15.6cm}{!}{
\begin{tcolorbox}[fontupper=\small, fontlower=\small,title=Interpolation localisation: \textit{predicate-is}]
\underline{source: the sun \textcolor{blue}{is} in the northern hemisphere}\\ \\
1. the sun \textcolor{blue}{is} located in the northern hemisphere \\
2. the sun \textcolor{blue}{is} in the northern hemisphere \\
3. the sun \textcolor{blue}{is} made of air around the sun \\
4. the sun \textcolor{blue}{is} a source of sunlight for organisms \\
5. the sun \textcolor{blue}{is} a source of sunlight for birds \\
6. the sun \textcolor{blue}{is} a source of energy for organisms living in an arctic environment \\
7. the sun \textcolor{blue}{is} a source of food for plants \\
8. food \textcolor{blue}{is} a source of oxygen ; water for plants \\
9. food \textcolor{blue}{is} a source of energy for plants by producing heat \\
10. food \textcolor{blue}{is} a source of energy for a plant or animal / living thing \\

1. the sun \textcolor{blue}{is} the dominant star in the night sky \\
2. the sun \textcolor{blue}{is} closer to the earth than it is to the sun \\
3. the sun \textcolor{blue}{is} a star in the night sky \\
4. the sun \textcolor{blue}{is} good for the environment by providing sunlight to plants \\
5. the atmosphere \textcolor{blue}{is} an environment for intensive farming \\ 
6. the respiratory system \textcolor{red}{carries} oxygen to the rest of the body \\
7. food \textcolor{red}{contains} nutrients ; water ; food energy \\
8. food \textcolor{blue}{is} the nutrient for ( plants ; animals ) \\
9. producers \textcolor{red}{are} a source of energy for producers by weathering \\
10. food \textcolor{blue}{is} a part of a plant / animals / living things \\

\underline{target: food \textcolor{blue}{is} a source of energy for animals / plants}
\end{tcolorbox}
% }
\caption{Interpolation examples (top: supervised INN, bottom: Optimus).}
\label{tab:guide_generation_app2}
\end{table*}

\begin{table*}[ht!]
\begin{tcolorbox}[fontupper=\small, fontlower=\small, title=Interpolation localisation: \textit{argument-animals} and \textit{predicate-require}]
\underline{source: \textcolor{blue}{animals require} food to survive}\\ \\
1. \textcolor{blue}{animals require} water to survive \\
2. \textcolor{blue}{animals require} food for survival \\
3. \textcolor{blue}{animals require} food for survival \\
4. \textcolor{blue}{animals require} nutrients from food \\
5. \textcolor{blue}{an animal requires} food for survival \\
6. \textcolor{blue}{an animal requires} food for survival \\
7. \textcolor{blue}{an animal requires} nutrients from producers \\
8. \textcolor{blue}{an animal requires} nutrients for survival \\
9. \textcolor{blue}{an animal requires} nutrients from food \\
10. \textcolor{blue}{an animal requires} nutrients from producers \\

1. \textcolor{blue}{animals} \textcolor{red}{need} sunglasses for protection  \\
2. \textcolor{blue}{animals} \textcolor{red}{live} in an environment \\
3. \textcolor{blue}{animals} \textcolor{red}{need} food to thrive \\
4. \textcolor{blue}{animals require} energy for survival \\
5. \textcolor{red}{a consumer} \textcolor{red}{uses} some of the food that is available \\
6. only \textcolor{red}{a producer} \textcolor{red}{eats} plants \\
7. \textcolor{red}{a human} \textcolor{red}{produces} its own food \\
8. \textcolor{blue}{an animal requires} nutrients in a source of food to survive \\
9. \textcolor{blue}{an animal requires} energy to perform photosynthesis \\
10. \textcolor{blue}{an animal requires} nutrients to grow \\ \\
\underline{target: \textcolor{blue}{an animal requires} nutrients from producers}
\end{tcolorbox}
\caption{Interpolation examples (top: supervised INN, bottom: Optimus).}
\label{tab:guide_generation_app1}
\end{table*}

\section{INNs: Explanation Reconstruction} \label{sec:rec_example}
% Table \ref{tab:unsup_rec_explain} shows some generated explanations from AutoEncoder and unsupervised INN. As we can see, they can reconstruct the explanations with good quality.
% \input{tables/rec_examples1}

Table \ref{tab:rec_explain} shows some reconstructed explanations from AutoEncoder, unsupervised INN, and supervised INN, respectively.
% \begin{table*}[ht!]
% \scriptsize
% \begin{center}
% \begin{tikzpicture}
% \node (table) [inner sep=.1pt] {

\begin{table*}[ht!]
    % \resizebox{\textwidth}{15mm}{
% \begin{tikzpicture}
    % \scriptsize
    \small
    \centering
\renewcommand\arraystretch{1.1}
    \begin{tabular}{p{3.6cm}p{3.6cm}p{3.6cm}p{3.6cm}}  \toprule
        \textbf{Augmented explanations} & \textbf{BERT-GPT2} & \textbf{unsupervised INN} & \textbf{supervised INN} \\ \hline
        a animal requires water for survival & a animal requires water for survival & a animal requires water for survival & a animal requires water for survival \\ \hline
        an animal requires a mate for survival & an animal requires a mate to reproduce & an animal requires a mate to reproduce & an animal requires a reproductive system for survival \\ \hline
        some animals sometimes hunt for prey & some animals prey on other animals & some animals sometimes catch prey & some animals sometimes hunt for prey \\ \hline
        an animal requires energy of its own to move & an animal requires energy from somewhere to move & an animal requires energy to move & an animal requires energy for movement \\ \hline
        an animal requires energy to run & an animal requires energy to run & an animal requires energy to run & an animal requires energy to run \\ \hline
        animals live in their habitats & animals live in their habitats & animals live in their habitat & animals live in their habitat \\ \hline
        animals must eat animals to survive & animals must eat to survive & animals must eat other animals to survive & animals must eat to survive \\ \hline
        animals taste flavors & animals taste flavors & animals taste flavors & animals taste flavors \\ \hline
        % some animals must catch prey to survive  & some animals may catch prey in their droppings & some animals must catch prey to survive & some animals must catch prey before they can reproduce \\ \hline
        animals eat plants & animals eat plants & animals eat plants & animals eat plants\\ \hline
        an animal requires nutrients to grow and heal & an animal requires nutrients in soil for survival & an animal requires nutrients to grow and repair & an animal needs to store fat to grow \\ \hline
        animals require oxygen to grow & animals require oxygen to grow & animals require oxygen to breath & animals require oxygen for survival \\ \hline
        an animal needs to breathe in order to survive &an animal requires food for survival & a animal needs to breathe to survive & an animal requires water and food to survive \\ \hline
        humans cause the disease & humans cause the disease & humans cause the disease & humans cause the disease \\ \hline
        humans have a negative impact on the environment & humans have a negative impact on the ecosystem & humans have a negative impact on the environment & humans have a negative impact on the environment \\ \hline
        humans require water to survive & humans require water to survive & humans require water for survival & humans require water for survival \\ \hline
        humans produce offspring & humans produce offspring & humans eat plants & humans produce offspring \\ \hline
        humans have lived on earth & humans live in the solar system & humans live in the solar system & humans live in the biosphere \\ \hline
        humans use fossil fuels for energy & humans use fossil fuels to make energy & humans use fossil fuels to make energy & humans use natural gas to make energy \\ \hline
        humans eat green plants & humans eat green plants & humans eat green plants & humans eat green plants \\ \hline
        humans eat fruit & humans eat fruit & humans eat fruit & humans eat fruit \\ \hline
        % humans may use coal to produce electricity & humans may use coal to make electrical energy & fossil fuels can be used to power cars  & producers usually use coal for power \\ \hline
        humans sometimes eat plants or animals & humans sometimes eat plants and animals & living things sometimes eat insects / animals & animals sometimes eat seeds from trees \\ \hline
        a plant absorbs light energy for photosynthesis & a plant absorbs sunlight for photosynthesis & an flower requires energy to grow and provide warmth to the skin & a plant absorbs light for photosynthesis \\ \hline
        % a plant uses carbohydrates to grow and thrive & a plant uses carbohydrates to make food for itself & a plant requires nutrients from soil for plant reproduction & a plant uses photosynthesis to make food for itself \\ \hline
        a plant absorbs water from the air into its roots & a plant absorbs water from the air into its body & a leaf absorbs water from the air through the leaves & a plant absorbs water and nutrients from the air \\ \hline
        a plant uses energy to grow & a plant requires energy for growth & a plant requires energy to grow & a plant requires energy to grow \\ \hline
        plant reproduction occurs in the spring & plant reproduction occurs in the spring & plant reproduction begins during seed dispersal & plant reproduction begins in spring \\ \hline
        plants require water and sunlight to grow & plants require water and sunlight to grow & plants require sunlight to grow and survive & plants require water and sunlight to grow \\ \hline
        a plant requires a habitat for survival & a plant needs a habitat for survival & a plant requires a habitat for survival & a plant requires a habitat for survival \\ \toprule
        
    \end{tabular}
    \caption{Explanation reconstruction. From left to right are augmented explanations, decoded explanations from AutoEncoder, explanations from unsupervised INN, and that from supervised INN, respectively.} 
    \label{tab:rec_explain}
\end{table*}

\end{document}